\documentclass{article}

\usepackage{microtype}
\usepackage{graphicx}
\usepackage{subfigure}
\usepackage{booktabs} 

\usepackage{hyperref}

\usepackage[accepted]{icml2020_mod}

\usepackage{graphicx}  
\usepackage{ amssymb }
\usepackage{amsmath}
\usepackage{bbm}
\usepackage{mathtools}
\usepackage{xcolor}

\usepackage{amssymb}
\usepackage{amsmath}
\usepackage{bbm}
\usepackage{mathtools}

\icmltitlerunning{Stiffness: A New Perspective on Generalization in Neural Networks}

\begin{document}
	
	\twocolumn[
	\icmltitle{Stiffness: A New Perspective on Generalization in Neural Networks}
	
	\icmlsetsymbol{equal}{*}
	
	\begin{icmlauthorlist}
		\icmlauthor{Stanislav Fort}{google,stanford,residency}
		\icmlauthor{Pawe\l{} Krzysztof Nowak}{google}
		\icmlauthor{Stanis\l{}aw Jastrzebski}{google,nyu,intern}
		\icmlauthor{Srini Narayanan}{google}
	\end{icmlauthorlist}
	
	\icmlaffiliation{stanford}{Stanford University, Stanford, CA, USA }
	\icmlaffiliation{google}{Google Research, Zurich, Switzerland}
	\icmlaffiliation{residency}{This work has been done as a part of the Google AI Residency program}
	\icmlaffiliation{intern}{A part of this work was done while an intern in Google Research, Zurich, Switzlerland}
	\icmlaffiliation{nyu}{New York University, New York, USA}
	
	\icmlcorrespondingauthor{Stanislav Fort}{sfort1@stanford.edu}
	
	\vskip 0.3in
	]
	
	\printAffiliationsAndNotice{} 
	
	\begin{abstract}
		In this paper we develop a new perspective on generalization of neural networks by proposing and investigating the concept of a neural network stiffness. We measure how stiff a network is by looking at how a small gradient step in the network’s parameters on one example affects the loss on another example. Higher stiffness suggests that a network is learning features that generalize. In particular, we study how stiffness depends on 1) class membership, 2) distance between data points in the input space, 3) training iteration, and 4) learning rate. We present experiments on MNIST, FASHION MNIST, and CIFAR-10/100 using fully-connected and convolutional neural networks, as well as on a transformer-based NLP model. We demonstrate the connection between stiffness and generalization, and observe its dependence on learning rate. When training on CIFAR-100, the stiffness matrix exhibits a coarse-grained behavior indicative of the model's awareness of super-class membership. In addition, we measure how stiffness between two data points depends on their mutual input-space distance, and establish the concept of a dynamical critical length -- a distance below which a parameter update based on a data point influences its neighbors.
	\end{abstract}
	
	\section{Introduction}
	Neural networks are a class of highly expressive function approximators that proved to be successful in many domains such as vision, natural language understanding, and game-play. The specific details that lead to their expressive power have recently been studied in \citet{Montfar2014OnTN,pmlr-v70-raghu17a,DBLP:conf/nips/PooleLRSG16}. Empirically, neural networks have been extremely successful at generalizing to new data despite their over-parametrization for the task at hand, as well as their  ability to fit arbitrary random data perfectly \citet{Zhang2016UnderstandingDL,closerlook}.
	\begin{figure}[!t]
		\centering
		\includegraphics[width=1.0\linewidth]{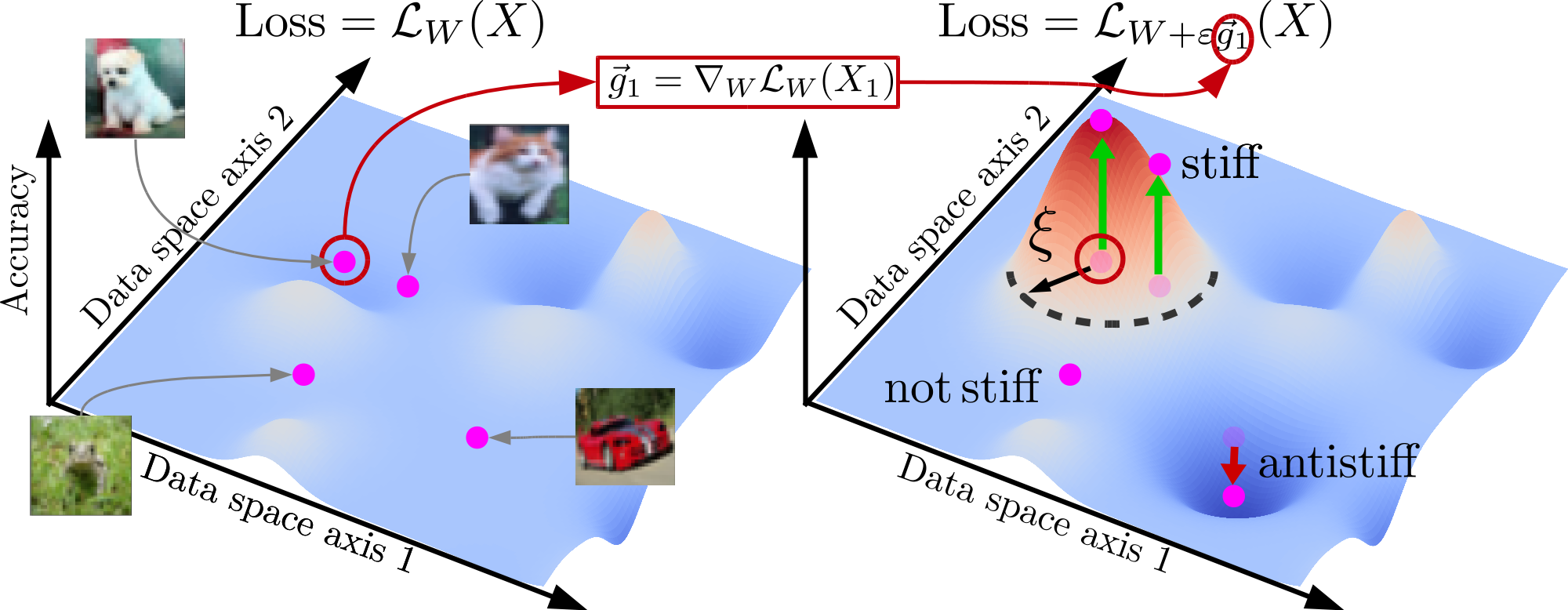}
		\caption{A diagram illustrating a) the concept of stiffness and b) dynamical critical length $\xi$. A small gradient update to the network's weights based on example $X_1$ decreases loss on some examples (stiff), and increases it on others (anti-stiff). We call the characteristic distance over which datapoints are stiff the dynamical critical length $\xi$.}
		\label{fig:3D_diagram}
	\end{figure}

	The fact that gradient descent is able to find good solutions given the highly over-parametrized family of functions has been studied theoretically in \citet{DBLP:journals/corr/abs-1802-06509} and explored empirically in \citet{DBLP:journals/corr/abs-1804-08838}, where the effective low-dimensional nature of many common learning problems is shown. \citet{goldilocks} extends the analysis in \citet{DBLP:journals/corr/abs-1804-08838} to demonstrate the role of initialization on the effective dimensionality, and \citet{fort2019large} use the result to build a phenomenological model of the loss landscape. 
	
	\citet{du_gradient_2018} and \citet{du_gradient_2018-1} use a Gram matrix to study convergence in neural network empirical loss. \citet{NIPS2017_6857} study the concentration properties of a similar covariance matrix formed from the output of the network. \citet{ghorbani2019investigation,jastrzebski2018on} investigate the Hessian eigenspectrum and \citet{Papyan2019MeasurementsOT} show how it is related to the gradient covariance. The clustering of logit gradients is studied e.g. in \citet{fort2019emergent}. Both concepts are closely related to our definition of stiffness.
	
	To explain the remarkable generalization properties of neural networks, it has been proposed \citep{spectralbias} that the function family is biased towards low-frequency functions. The role of similarity between the neural network outputs to similar inputs has been studied in \citet{DBLP:journals/corr/SchoenholzGGS16} for random initializations and explored empirically in \citet{DBLP:journals/corr/abs-1802-08760}, and it plays a crucial role in the Neural Tangent Kernel work started by \citet{jacot2018neural}. 
	
	\subsection{Our contribution}
	In this paper, we study generalization through the lens of \textit{stiffness}. We measure how \textit{stiff} a neural network is by analyzing how a small gradient step based on one input example affects the loss on another input example. Stiffness captures the resistance of the functional approximation learned to deformation by gradient steps. 
	
	We find the concept of stiffness useful 1) in diagnosing and characterizing generalization, 2) in uncovering semantically meaningful groups of datapoints beyond classes, 3) exploring the effect of learning rate on the function learned, and 4) defining and measuring the \textit{dynamical critical length} $\xi$. We show that higher learning rates bias the functions learned towards lower, more local stiffness, making them easier to bend on gradient updates. We demonstrate these on vision tasks (MNIST, FASHION MNIST, CIFAR-10/100) with fully-connected and convolution architectures (including ResNets), and on a natural language task using a fine-tuned BERT model.
	
	We find stiffness to be a useful concept to study because it a) relates directly to generalization, b) it is sensitive to semantically meaningful content of the inputs, and c) at the same time relates to the loss landscape Hessian and therefore to the research on its curvature and optimization. As such, it bridges the gap between multiple directions simultaneously.
	
	This paper is structured as follows: we introduce the concept of stiffness and the relevant theory in Section~\ref{sec:theory}. We describe our experimental setup in Section~\ref{methods}, present their results in Section~\ref{sec:results_and_discussion}, and discuss their implications in Section~\ref{sec:discussion}. We conclude wit Section~\ref{sec:conclusion}.
	
	\section{Theoretical background}
	\label{sec:theory}
	\subsection{Stiffness -- definitions}
	Let a functional approximation (e.g. a neural network) $f$ be parametrized by tunable parameters $W$. Let us assume a classification task and let a data point $X$ have the ground truth label $y$. A loss $\mathcal{L}(f_W(X),y)$ gives us the amount of mismatch between the function's output at input $X$ and the ground truth $y$. The gradient of the loss with respect to the parameters
	\begin{equation}
		\vec{g} = \nabla_W \mathcal{L}(f_W(X),y) \,
	\end{equation}
	is the direction in which, if we were to change the parameters $W$, the loss would change the most rapidly (for infinitesimal step sizes). Gradient descent uses this step (the negative of it) to update the weights and gradually tune the functional approximation to better correspond to the desired outputs on the training dataset inputs.
	\begin{figure}[!h]
		\centering
		\includegraphics[width=1.0\linewidth]{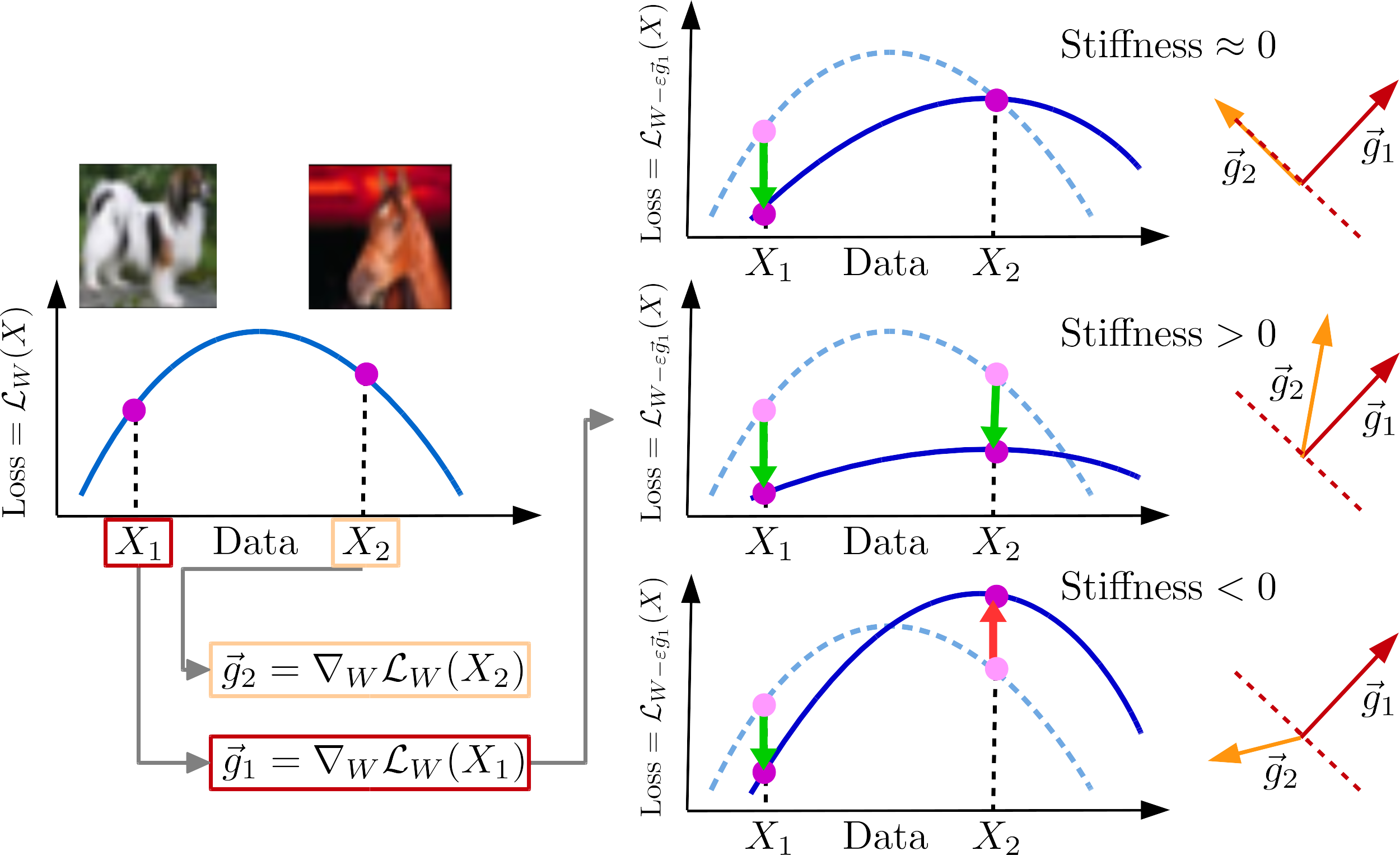}
		\caption{A diagram illustrating the concept of stiffness. It can be viewed in two equivalent ways: a) as the change in loss at a datapoint induced by the application of a gradient update based on another datapoint, and b) the alignment of loss gradients computed at the two datapoints. These two descriptions are mathematically equivalent.}
		\label{fig:diagram}
	\end{figure}
	Let us consider two data points with their ground truth labels $(X_1,y_1)$ and $(X_2,y_2)$. We construct a gradient with respect to example 1 as $\vec{g}_1 = \nabla_W \mathcal{L}(f_W(X_1),y_1)$ and ask how do the losses on data points 1 and 2 change as the result of a small change of $W$ in the direction $-\vec{g}_1$, i.e. what is 
	\begin{equation}
		\Delta \mathcal{L}_1 = \mathcal{L} (f_{W-\varepsilon \vec{g}_1}(X_1),y_1) - \mathcal{L} (f_{W}(X_1),y_1)\, ,
	\end{equation}
	which is equivalent to 
	\begin{equation}
		\Delta \mathcal{L}_1 = -\varepsilon \nabla_\varepsilon \mathcal{L}(f_{W-\varepsilon \vec{g}_1}(X_1),y_1) = - \varepsilon \vec{g}_1 \cdot \vec{g}_1 + \mathcal{O}(\varepsilon^2)\, .
	\end{equation}
	The change in loss on input 2 due to the same gradient step from input 1 becomes equivalently $\Delta \mathcal{L}_2 = - \varepsilon \nabla_\varepsilon \mathcal{L}(f_{W-\varepsilon \vec{g}_1}(X_2),y_2) = - \varepsilon \vec{g}_1 \cdot \vec{g}_2 + \mathcal{O}(\varepsilon^2)$.
	We are interested in the correlation in loss changes $\Delta \mathcal{L}_1$ and $\Delta \mathcal{L}_2$, and will work in the limit of $\varepsilon \to 0$, i.e. an infinitesimal step size. We know that $\Delta \mathcal{L}_1 < 0$ since we constructed the gradient update accordingly. We define positive stiffness to mean $\Delta \mathcal{L}_2 < 0$ as well, i.e. that losses at both inputs went down. We assign the stiffness of 0 for $\Delta \mathcal{L}_2 = 0$. If $\Delta \mathcal{L}_2 > 0$, the two inputs would be anti-stiff (negative stiffness). The equations above show that this can equivalently be thought of as the overlap between the two gradients $\vec{g}_1 \cdot \vec{g}_2$ being positive for positive stiffness, and negative for negative stiffness. We illustrate this in Figure~\ref{fig:3D_diagram} and Figure~\ref{fig:diagram}.
	
	The above indicate that what we initially conceived of as a change in loss due to the application of a small gradient update from one input to another is in fact equivalent to analyzing \emph{gradient alignment} between different datapoints.

	We will be using 2 different definitions of stiffness: the sign stiffness and the cosine stiffness. We define the \textit{sign} stiffness to be the expected sign of $\vec{g_1} \cdot \vec{g_2}$ (or equivalently the expected sign of $\Delta \mathcal{L}_1 \Delta \mathcal{L}_2$)  as
	\begin{equation}
		S_\mathrm{sign}((X_1,y_1),(X_2,y_2); f) = \mathop{\mathbb{E}} \left [ \mathrm{sign} \left ( \vec{g_1} \cdot \vec{g_2} \right ) \right ] \, ,
	\end{equation}
	where stiffness depends on the dataset from which $X_1$ and $X_2$ are drawn. The cosine stiffness is
	\begin{equation}
		S_\mathrm{cos}((X_1,y_1),(X_2,y_2); f) = \mathop{\mathbb{E}} \left [ \mathrm{cos} \left ( \vec{g_1} , \vec{g_2} \right ) \right ] \, ,
	\end{equation}
	where $\mathrm{cos} \left ( \vec{g_1}, \vec{g_2} \right ) = (\vec{g_1} / |\vec{g_1}|) \cdot (\vec{g_2} / |\vec{g_2}|)$. We use both versions of stiffness as they are suitable to highlight different phenomena -- the \textit{sign} stiffness shows the stiffness between classes clearer, while the \textit{cosines} stiffness is more useful for within-class stiffness.  
	
	Based on our definitions, the stiffness between a datapoint $X$ and itself is 1.

	\subsection{Train-train, train-val, and val-val}
	When measuring stiffness between two datapoints, we have 3 options: 1) choosing both datapoints from the training set (we call this \textit{train-train}), 2) choosing one from the training set and the other from the validation set (\textit{train-val}), and 3) choosing both from the validation set (\textit{val-val}). The train-val stiffness is \textit{directly} related to generalization, as it corresponds to the amount of improvement on the training set transferring to the improvement of the validation set. We empirical observe that all 3 options are remarkably similar, which gives us confidence that they all are related to generalization. This is further supported by observing their behavior is a function of epoch in Figures~\ref{fig:overfitting}, \ref{fig:CIFAR100_resnet_hierarchy}, and \ref{fig:BERT_MNLI}.
	
	\subsection{Stiffness based on class membership}
	A natural question to ask is whether a gradient taken with respect to an input $X_1$ in class $c_1$ will also decrease the loss for example $X_2$ in class $c_2$. In particular, we define the \textit{class stiffness matrix} 
	\begin{equation}
		C(c_a,c_b) = \mathop{\mathbb{E}}_{X_1 \in c_a, X_2 \in c_b, X_1 \neq X_2} \left [ S((X_1,y_1),(X_2,y_2))  \right ] \, .
	\end{equation}
	The on-diagonal elements of this matrix correspond to the suitability of the current gradient update to the members of the class itself. In particular, they correspond to \textit{within class} generalizability (we avoid counting the stiffness of an example to itself, as that is by definition 1). The off-diagonal elements, on the other hand, express the amount of improvement transferred from one class to another. They therefore directly diagnose the amount of generality the currently improved features have.
	
	A consistent summary of generalization between classes is the off-diagonal sum of the class stiffness matrix
	\begin{equation}
		S_{\text{between classes}} = \frac{1}{N_c (N_c -1)} \sum_{c_1} \sum_{c_2 \neq c_1} C(c_1,c_2) \, . 
	\end{equation}
	In our experiments, we track this value as a function of learning rate once we reached a fixed loss. The quantity is related to how generally applicable the learned features are, i.e. how well they transfer from one class to another. For example, for CNNs learning good edge detectors in initial layers typically benefits all downstream tasks, regardless of the particular class in question. We do the equivalent for the within-class stiffness (= on diagonal elements). When the within-class stiffness starts going $<1$, the generality of the features improved does not extend to even the class itself, suggesting an onset of overfitting (see Figure~\ref{fig:overfitting}).
	
	\subsection{Higher order classes}
	For some datasets (in our case CIFAR-100), there exist large, semantically meaningful \textit{super}-classes to which datapoints belong. We observe how stiffness depends on the datapoints' membership in those super-classes, and going a step further, define groups of super-classes we call super-super-classes. We observe that the stiffness within those semantically meaningful groups is higher than would be expected on average, suggesting that stiffness is a measure is aware of the semantic content of the datapoints (Figure~\ref{fig:CIFAR100_resnet_superclasses}).
	
	\subsection{Stiffness as a function of distance}
	We investigate how stiff two datapoints are, on average, as a function of how far away from each other they are in the input space. We can think of neural networks as a form of kernel learning (in some limits, this is exact, e.g. \citet{jacot2018neural}) and here we are investigating the particular form of the kernel learned, in particular its direction-averaged sensitivity. This links our results to the work on spectral bias (towards slowly-varying, low frequency functions) in \citet{spectralbias}. We are able to directly measure the characteristic size of the stiff regions in neural networks trained on real tasks which we call \textit{dynamical critical length} $\xi$. The concept is illustrated in Figure~\ref{fig:3D_diagram}. We work with data normalized to the unit sphere $|\vec{X}| = 1$ and use their mutual cosine to define their distance as
	\begin{equation}
		\mathrm{distance}(\vec{X}_1,\vec{X}_2) = 1 - \frac{\vec{X}_1 \cdot \vec{X}_2}{|\vec{X}_1||\vec{X}_2|} \, .
	\end{equation}
	This has the advantage of being bounded between 0 and 2. For $X_1 = X_2$, i.e. $\mathrm{distance}=0$, stiffness is 1 by definition. We track the threshold distance $\xi$ at which stiffness, on average, goes to 0. We observe $\xi$ as a function of training and learning rate to estimate the characteristic size of the stiff regions of a neural net.
	
	\section{Methods}
	\label{methods}

	\begin{figure}[htbp]
		\includegraphics[width=0.9\linewidth]{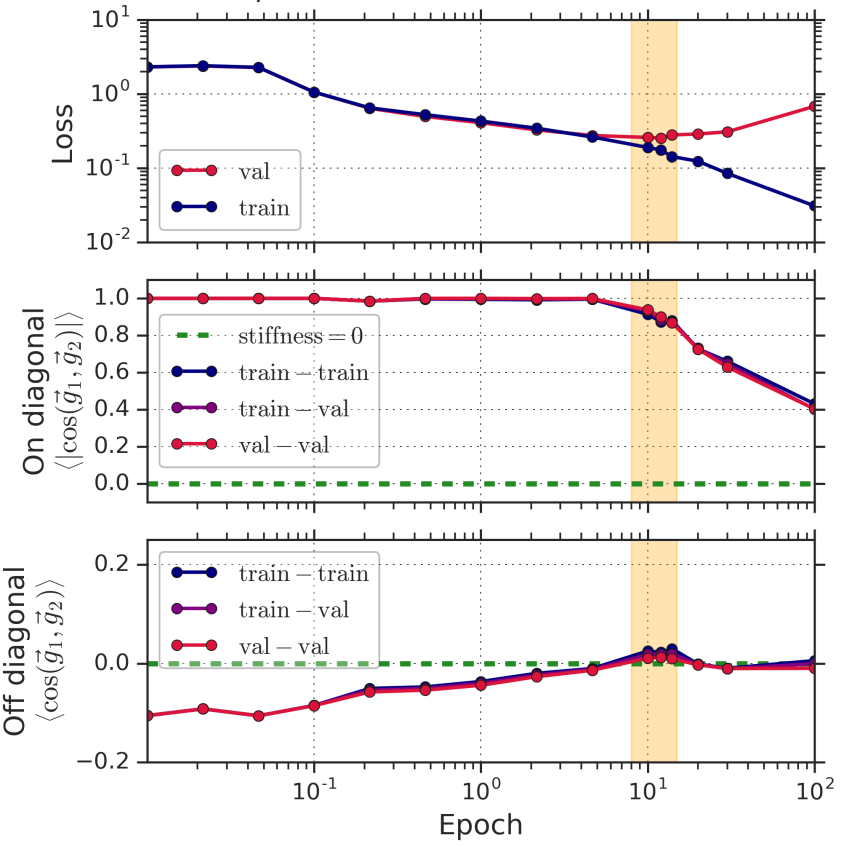}
		\caption{The evolution of training and validation loss (\textit{top panel}), within-class stiffness (\textit{central panel}) and between-classes stiffness (\textit{bottom panel}) during training. The onset of over-fitting is marked in orange. After that, both within-class and between-classes stiffness regress to 0. The same effect is visible in stiffness measured between two training set datapoints, one training and one validation datapoint, and two validation set datapoints.}
		\label{fig:overfitting}
	\end{figure}
	\subsection{Experimental setup}
	We ran a large number of experiments with fully-connected (FC) and convolutional neural networks (CNN) (both small, custom-made networks, as well as ResNet \citet{he2015deep}) on 4 classification datasets: MNIST \citep{lecun-mnisthandwrittendigit-2010}, FASHION MNIST \citet{DBLP:journals/corr/abs-1708-07747}, CIFAR-10 \citet{Krizhevsky2009LearningML}, and CIFAR-100 \citet{Krizhevsky2009LearningML}. In addition, we study the behavior of a BERT \citet{BERT} fine-tuned on the MNLI dataset \citet{MNLI}.
	
	Using those experiments, we investigated the behavior of stiffness as a function of 1) training epoch, 2) the choice of learning rate, 3) class membership, and 4) the input space distance between images.
	
	For experiments with fully-connected neural networks, we used a 3 layer $\mathrm{ReLU}$ network of the form $X \to 500 \to 300 \to 100 \to y$. For experiments with convolutional neural networks, we used a 3 layer network with filter size 3 and the numbers of channels being 16, 32, and 32 after the respective convolutional layer, each followed by $2 \times 2$ max pooling. The final layer was fully-connected. For ResNets, we used the ResNet20v1 implementation from \citet{chollet2015keras}. No batch normalization was used.
	
	We pre-processed the network inputs to have zero mean and unit variance, and normalized all data to the unit sphere as $|\vec{X}| = 1$. We used \begin{math} \mathrm{Adam} \end{math} with different (constant) learning rates as our optimizer and the default batch size of 32. All experiments took at most a few hours on a single GPU.
	
	\subsection{Training and stiffness evaluation}
	We trained a network on a classification task first, to the stage of training at which we wanted to measure stiffness. We then froze the weights and evaluated loss gradients on a the relevant set of images.
	
	We evaluated stiffness properties between pairs of data points from the training set, one datapoint from the training and one from the validation set, and pairs of points from the validation set. We used the full training set to train our model. The procedure was as follows: 1) Train for a number of steps on the \textit{training} set and update the network weights accordingly. 2) For each of the modes \{ train-train, train-val, and val-val\}, go through tuples of images coming from the respective datasets. 3) For each tuple calculate the loss gradients $g_1$ and $g_2$, and compute check $\mathrm{sign}(\vec{g}_1 \cdot \vec{g}_2)$ and $\mathrm{cos}(\vec{g}_1,\vec{g}_2)$. 4) Log the input space distance between the images as well as other relevant features.
	In our experiments, we used a fixed subset (typically of $\approx 500$ images for experiments with 10 classes, and $\approx 3,000$ images for 100 classes) of the training and validation sets to evaluate the stiffness properties on. We convinced ourselves that such a subset is sufficiently large to provide measurements with small enough statistical uncertainties, which we overlay in our figures.
	
	\subsection{Learning rate dependence}
	We investigated how stiffness properties depend on the learning rate used in training. To be able to do that, we first looked at the dynamical critical scale $\xi$ for different stages of training of our network, and then compared those based on the epoch and training loss at the time in order to be able to compare training runs with different learning rates fairly. The results are shown in Figures~\ref{fig:stiffness_vs_epoch} and \ref{fig:domain_size}, concluding that higher learning rates learn functions with more localized stiffness response, i.e. lower $\xi$.
	
	\section{Results}
	\label{sec:results_and_discussion}
	We will summarize our key results here and will provide additional figures in the Supplementary material.
	\subsection{Stiffness properties based on class membership}
	\begin{figure}[htbp]
		\centering
		\includegraphics[width=1.0\linewidth]{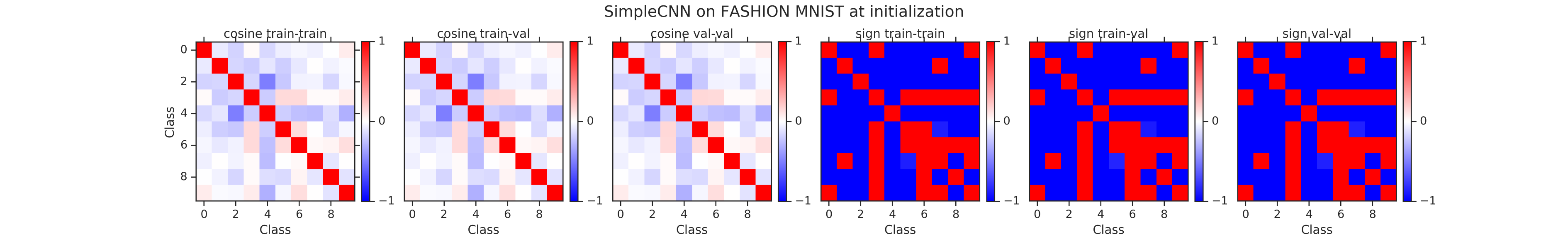}
		\includegraphics[width=1.0\linewidth]{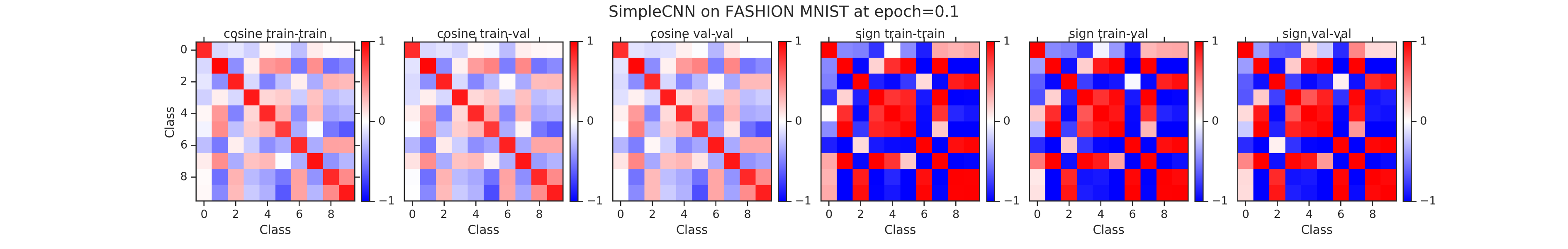}
		\includegraphics[width=1.0\linewidth]{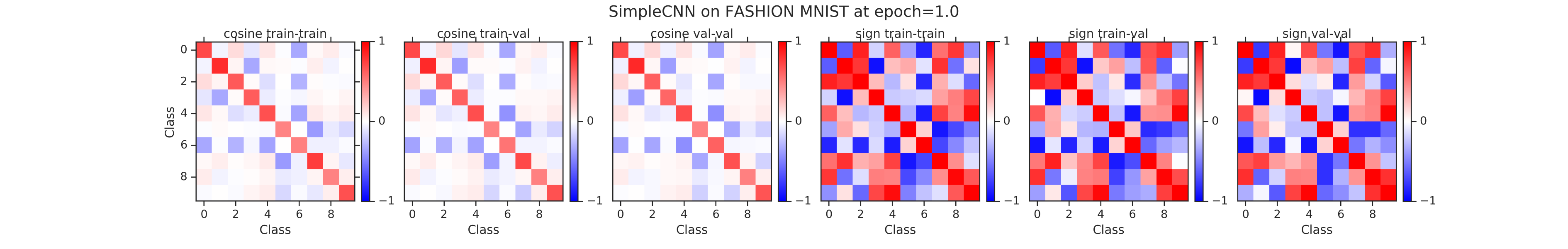}
		\includegraphics[width=1.0\linewidth]{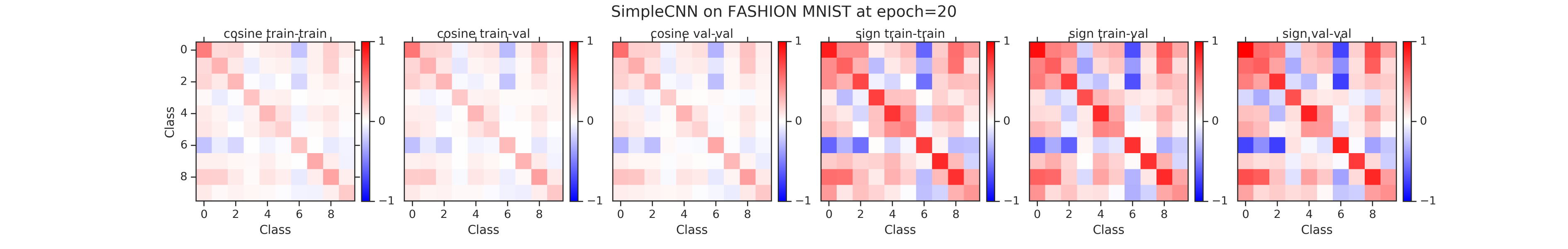}
		\caption{Class membership dependence of stiffness for a CNN on FASHION MNIST at 4 different stages of training. The figure shows stiffness between train-train, train-val and val-val pairs of images, as well as the sign and cosine metrics.}
		\label{fig:class_stiffness_CNN_FASHION_MNIST}
	\end{figure}
\begin{figure}[!h]
		\centering
		\includegraphics[width=1.0\linewidth]{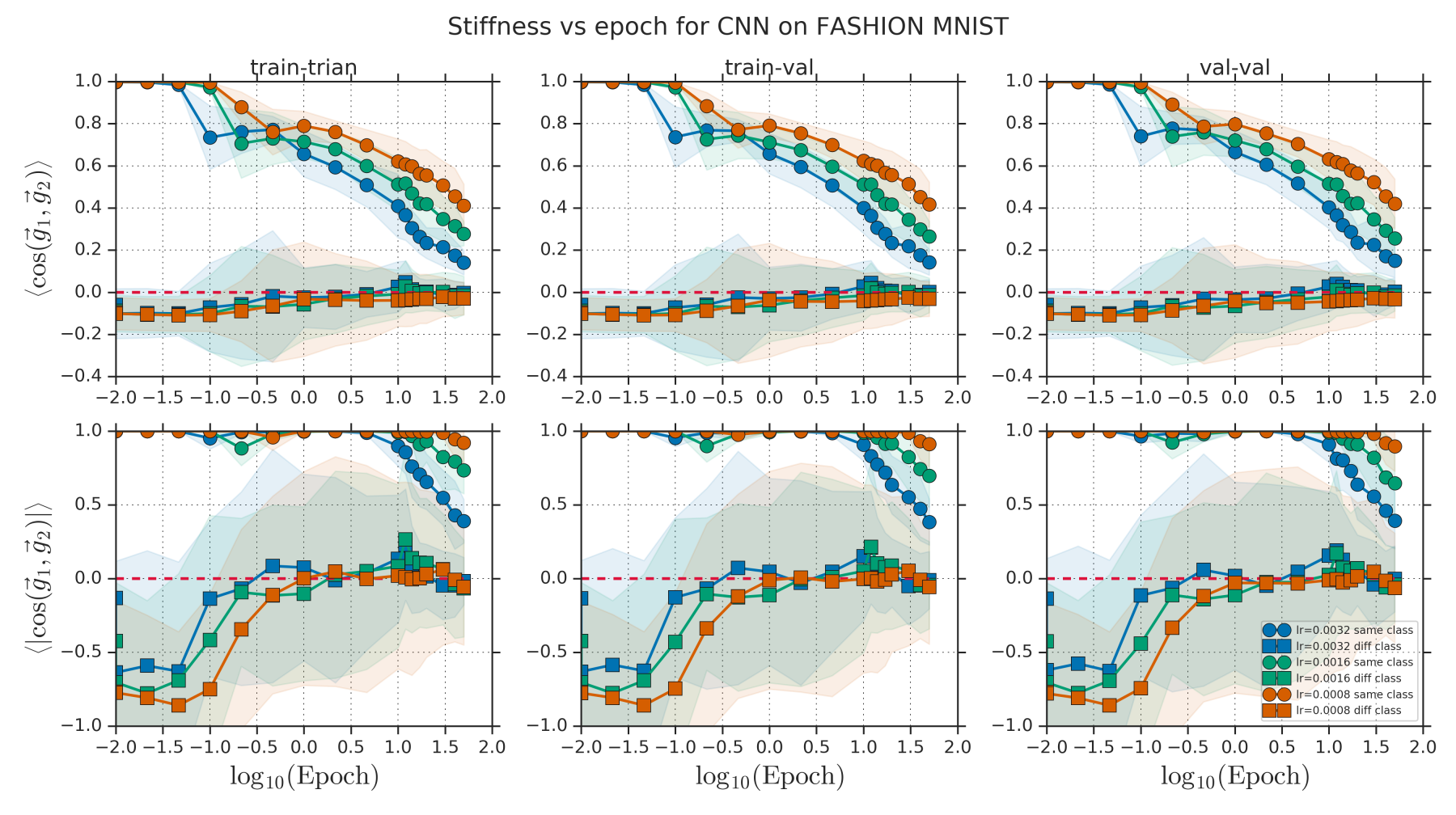} 
		\caption{Stiffness as a function of epoch. The plots summarize the evolution of within-class and between-classes stiffness measures as a function of epoch of training for a CNN on FASHION MNIST.}
		\label{fig:stiffness_vs_epoch}
	\end{figure}
\begin{figure}[!h]
		\centering
		\includegraphics[width=0.49\linewidth]{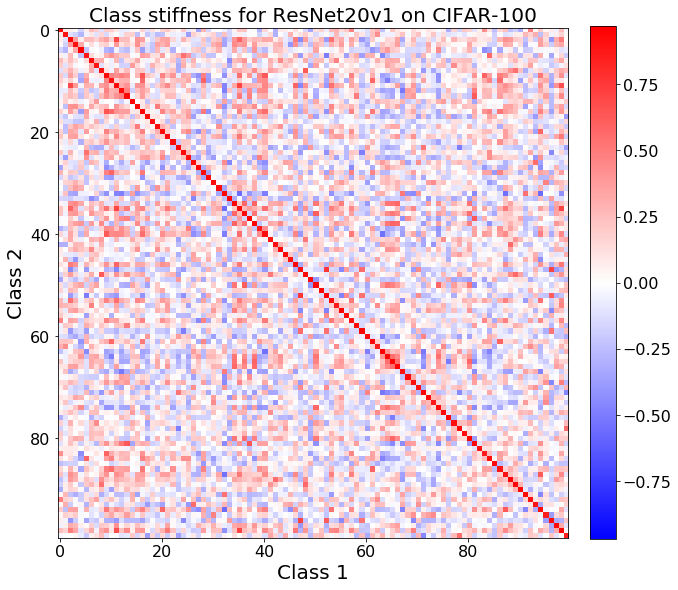}
		\includegraphics[width=0.49\linewidth]{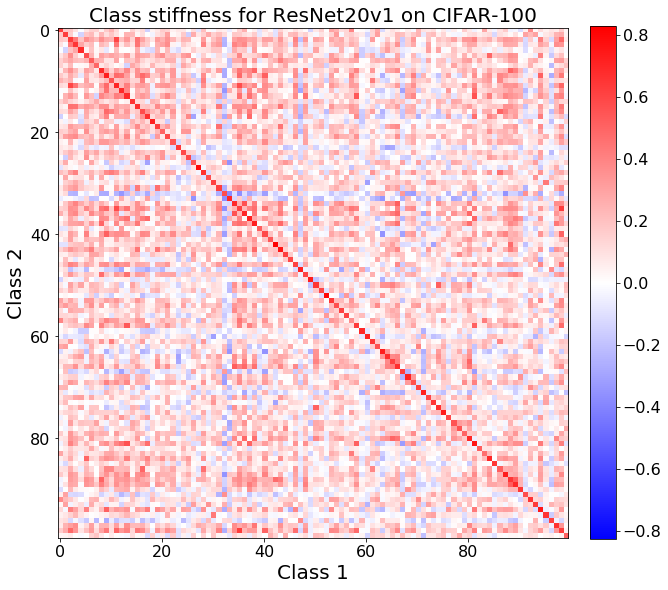}
		\includegraphics[width=0.49\linewidth]{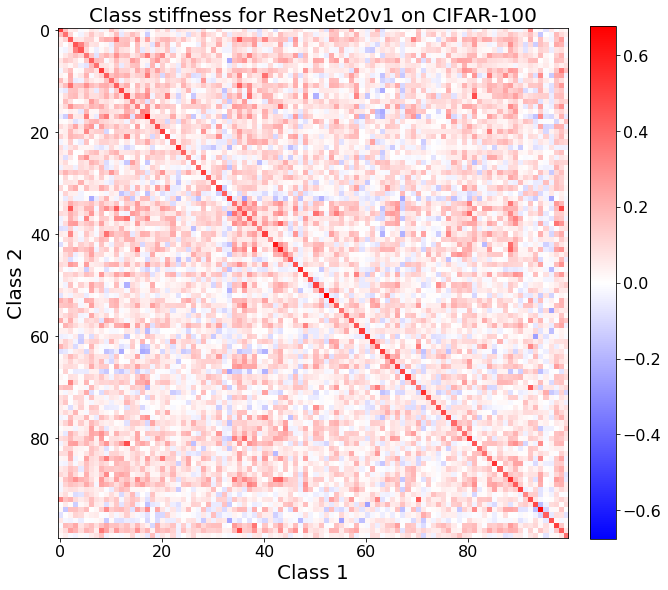}
		\includegraphics[width=0.49\linewidth]{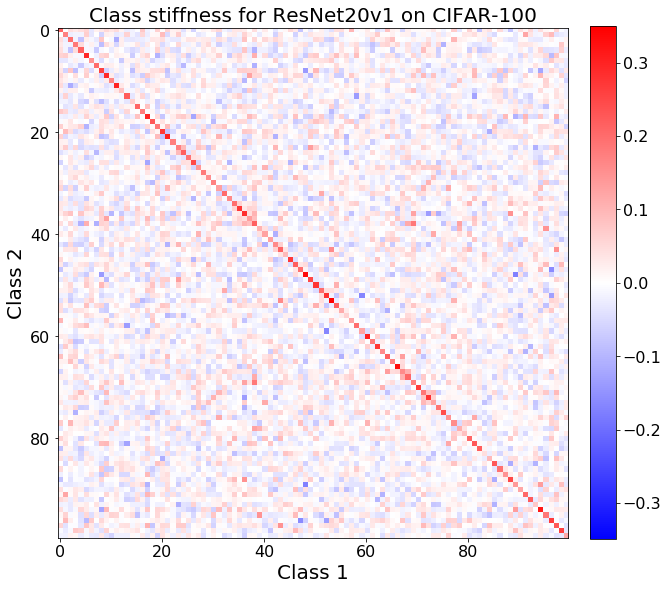}
		\caption{Stiffness between images of all pairs of classes for a ResNet20v1 trained on CIFAR-100. The within-class stiffness (the diagonal) is noticeable higher than the rest. The four panels show the evolution of the matrix through training, starting at upper left corner, and ending at the lower right corner.}
		\label{fig:CIFAR100_resnet_classes_evolution}
	\end{figure}

	We explored the stiffness properties based on class membership as a function of training epoch at different stages of training. Figure~\ref{fig:class_stiffness_CNN_FASHION_MNIST} shows results for a CNN on FASHION MNIST at initialization, very early in training, around epoch 1, and at the late stage. We provide additional results in the Supplementary material. The within-class (on diagonal) and between-classes results are summarized in Figure~\ref{fig:stiffness_vs_epoch}.
	Initially, an improvement based on an input from a particular class benefits only members of the same class. Intuitively, this could be due to some crude features shared within a class (such as the typical overall intensity, or the average color) being learned. There is no consistent stiffness between different classes at initialization. As training progresses, within-class stiffness stays high. In addition, stiffness between classes increases as well, given the model is powerful enough to learn the dataset. With the onset of overfitting, as shown in Figure~\ref{fig:overfitting}, the model becomes increasingly less stiff until even stiffness for inputs within the same class is lost. At that stage, an improvement on one example does not lead to a consistent improvement on other examples, as ever more detailed, specific features have to be learned. Figure~\ref{fig:CIFAR100_resnet_classes_evolution} shows the class-class stiffness matrix at 4 stages of training for a ResNet20v1 on CIFAR-100.
	\subsection{Class, super-class, and super-super-class structure of stiffness on CIFAR-100}
\begin{figure}[!h]
		\centering
		\includegraphics[width=0.75\linewidth]{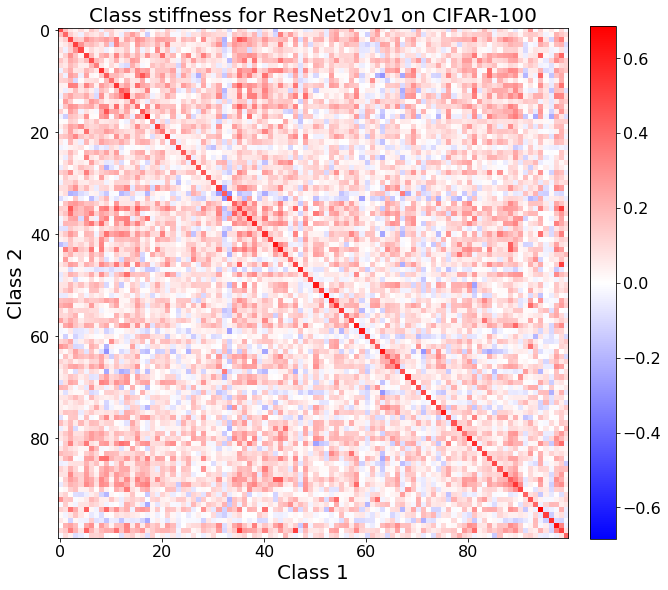}
		\caption{Stiffness between images of all pairs of classes for a ResNet20v1 trained on CIFAR-100. The within-class stiffness (the diagonal) is noticeable higher than the rest.}
		\label{fig:CIFAR100_resnet_classes}
	\end{figure}

	We studied the class-class stiffness between images from the CIFAR-100 dataset for a ResNet20v1 trained on the fine labels (100 classes) to approximately 60\% (top-1) test accuracy. As expected, the class-class stiffness was high for images from the same class (excluding the stiffness between the image and itself to prevent artificially increasing the value), as shown in Figure~\ref{fig:CIFAR100_resnet_classes}.
	
\begin{figure}[!h]
		\centering
		\includegraphics[width=1.0\linewidth]{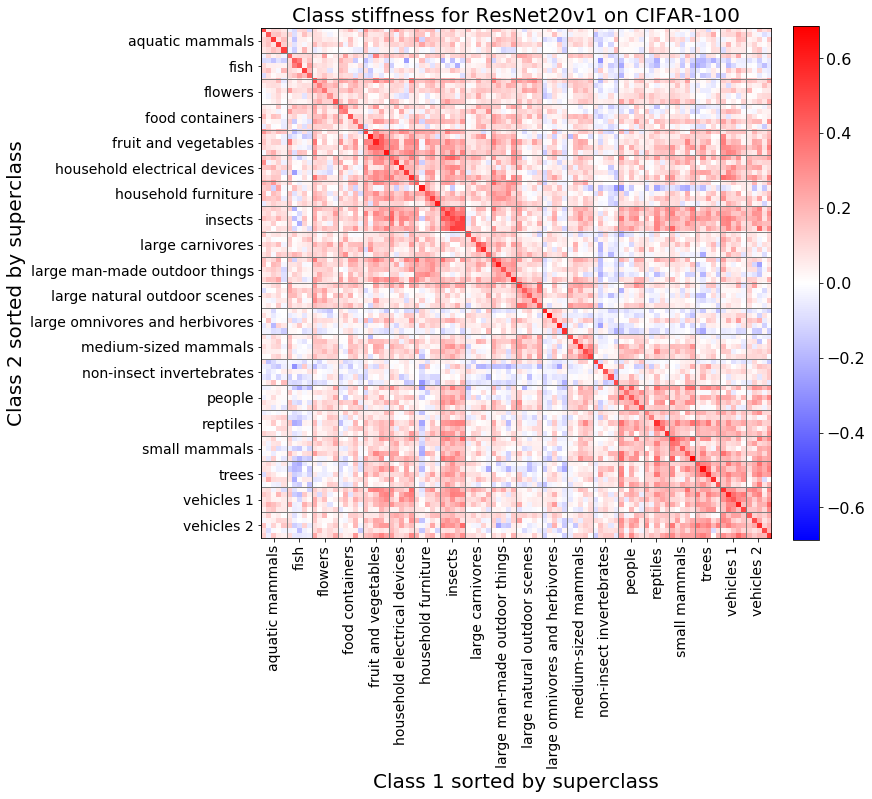}
		\caption{Stiffness between images of all pairs of classes for a ResNet20v1 trained on CIFAR-100, with classes sorted based on their super-class membership. The higher stiffness between classes from the same super-class is visible, indicating the sensitivity of stiffness to the semantically meaningful content of images. A higher order groups of super-classes are also showing.}
		\label{fig:CIFAR100_resnet_superclasses}
	\end{figure}

	We sorted the classes according to their super-classes (groups of 5 semantically connected classes) and visualized them in Figure~\ref{fig:CIFAR100_resnet_superclasses}. It seems that the groups of classes belonging to the same super-class are mutually stiff, without explicitly training the network to classify according to the super-class. Stiffness within the images of the same \textit{super-class} is noticeably higher than the typical value, and while it is not as high as within-class stiffness itself, this revelas that the network is aware of higher-order semantically meaningful categories to which the images belong. 
	Visually inspecting Figure~\ref{fig:CIFAR100_resnet_superclasses}, we see that there is an additional structure beyond the super-classes, which roughly corresponds to the \textit{living} / \textit{non-living} distinction.
	
	To quantify this hierarchy beyond a visual inspection of Figures~\ref{fig:CIFAR100_resnet_classes} and \ref{fig:CIFAR100_resnet_superclasses}, we calculated the following 4 measures: 1) the average stiffness between two different images from the same class, 2) the average stiffness between two different images from two different classes from the same super-class, 3) the average stiffness between two different images from two different classes from two different super-classes, belonging to the same super-super-class that we defined as either \textit{living} or \textit{non-living}, and 4) the average stiffness between examples from different classes, which served as a baseline. Each of these averages measures the amount of semantically meaningful similarity between reaction of images to gradient updates on gradually more abstract levels.
	\begin{figure}[!h]
		\centering
		\includegraphics[width=1.0\linewidth]{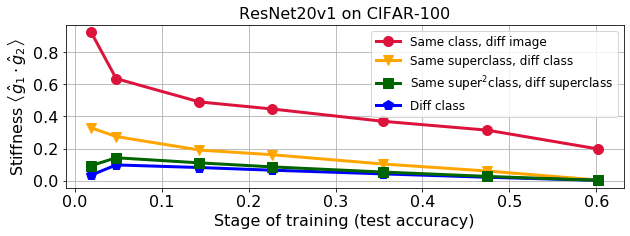}
		\caption{Stiffness between examples of the same class, super-class, and super-super-class for ResNet20v1 trained on the 100 fine classes of CIFAR-100. With an increasing epoch, all measures generically regress to 0. The yellow and green lines, corresponding to super-, and super-super-classes exceed the baseline (green), suggesting that the network is aware of higher-order semantic categories of the images throughout training.}
		\label{fig:CIFAR100_resnet_hierarchy}
	\end{figure}
	
	\subsection{Natural language tasks}
\begin{figure}[!h]
		\centering
		\includegraphics[width=1.0\linewidth]{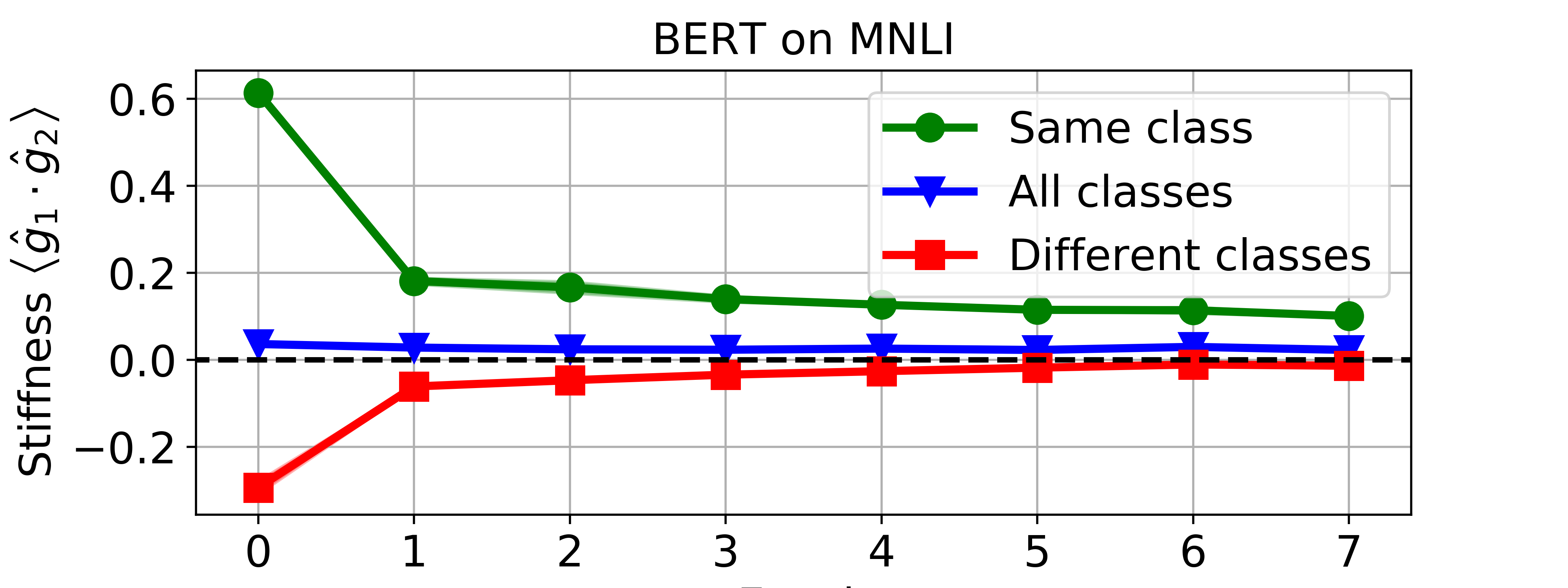}
		\caption{Stiffness as as function of epoch for a natural language task MNLI using a fine-tuned BERT model. Stiffness within the same class starts high and regresses towards 0, while stiffness between classes starts negative and approaches 0 from below. This is similar to our observations on vision tasks.}
		\label{fig:BERT_MNLI}
	\end{figure}

	To go beyond visual classification tasks, we studied the behavior of a BERT model fine tuned on the MNLI dataset. The evolution of within-class and between-classes stiffness is shown in Figure~\ref{fig:BERT_MNLI}. 
	It is very similar to our results on vision tasks such as Figure~\ref{fig:class_stiffness_CNN_FASHION_MNIST} and Figure~\ref{fig:CIFAR100_resnet_hierarchy}.
	
	\subsection{Stiffness as a function of distance between datapoints}
	\begin{figure}[!h]
		\centering
		\includegraphics[width=1.0\linewidth]{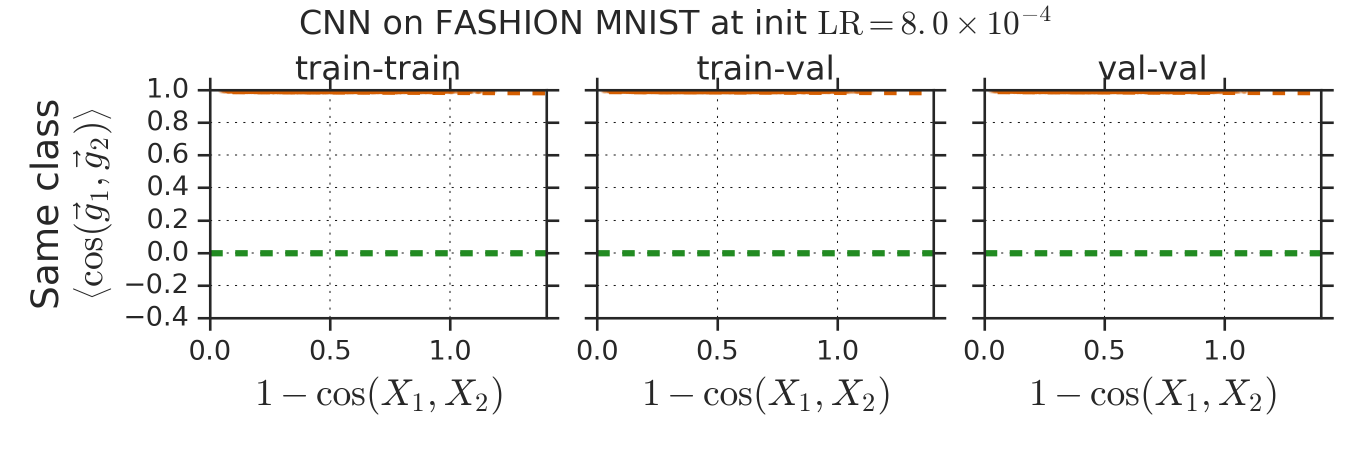}
		\includegraphics[width=1.0\linewidth]{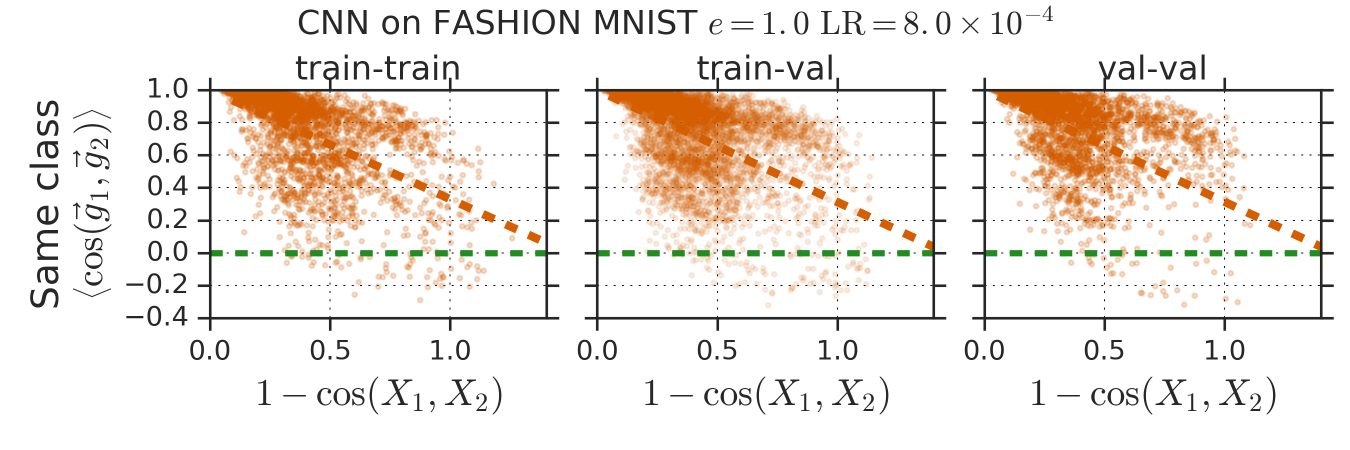}\\
		\includegraphics[width=1.0\linewidth]{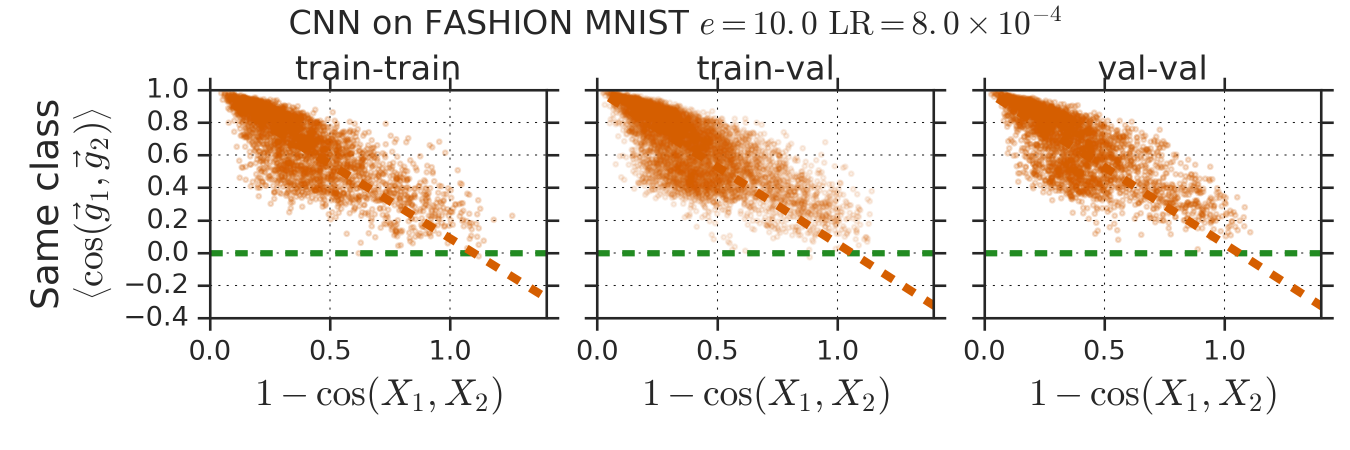}
		\includegraphics[width=1.0\linewidth]{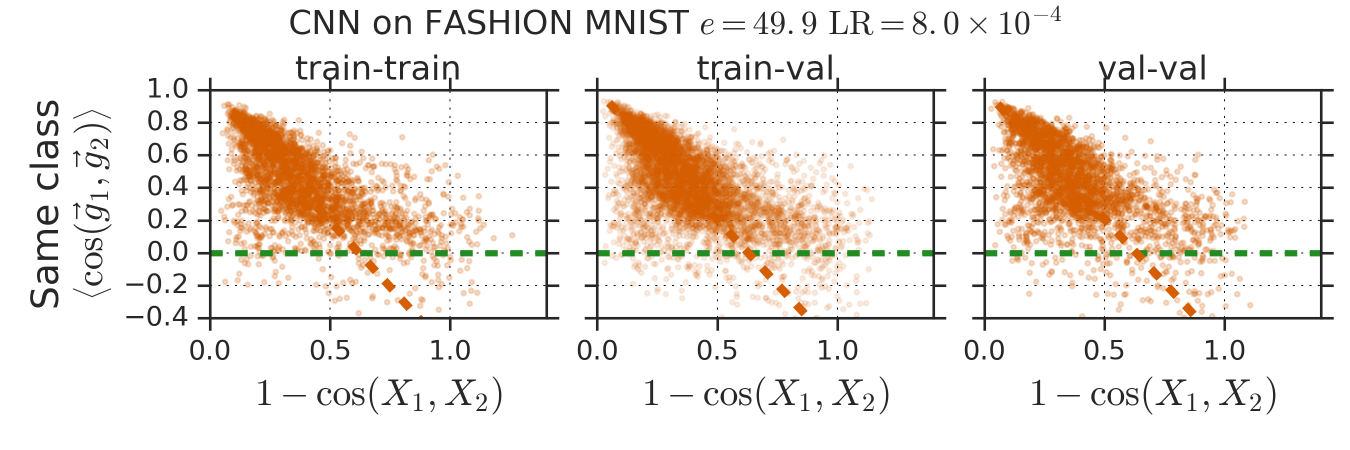}
		\caption{Stiffness between images of the same class as a function of their input space distance for 4 different stages of training of a CNN on FASHION MNIST. We add a linear fit to the data and measure its intersection with the $y=0$ line of 0 stiffness to estimate the dynamical critical length $\xi$. $\xi$ decreases with training time, suggesting that gradient updates influence distance datapoint less and less as training progresses, leading to eventually overfitting.}
		\label{fig:distance_dependence_CNN_FASHIONMNIST}
	\end{figure}
	We investigated stiffness between two datapoints as a function of their distance in the input space in order to measure how large the patches of the learned function that move together under gradient updates are (as illustrated in Figure~\ref{fig:3D_diagram}). We focused on datapoints from the same class. Examples of our results are shown in Figure~\ref{fig:distance_dependence_CNN_FASHIONMNIST} and in the Supplementary material. Fitting a linear function to the stiffness vs distance data, we estimate the typical distance at which stiffness goes to 0 for the first time, and called it the \textit{dynamical critical length} $\xi$. For a pair of identical datapoints, the stiffness is by definition 1. Equivalent plots were generated for each epoch of training in each of our experiments in order to measure $\xi$ used in Figure~\ref{fig:domain_size}.
	
	\subsection{The dynamical critical length $\xi$, and the role of learning rate}
	\begin{figure}[!h]
		\centering
		\includegraphics[width=1.0\linewidth]{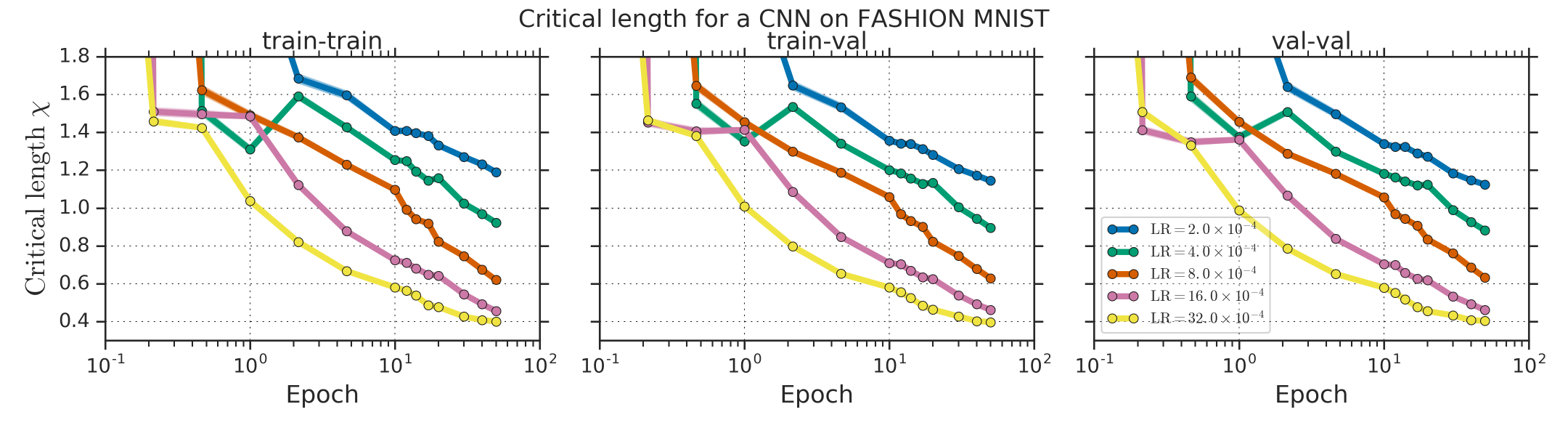} 
		\includegraphics[width=1.0\linewidth]{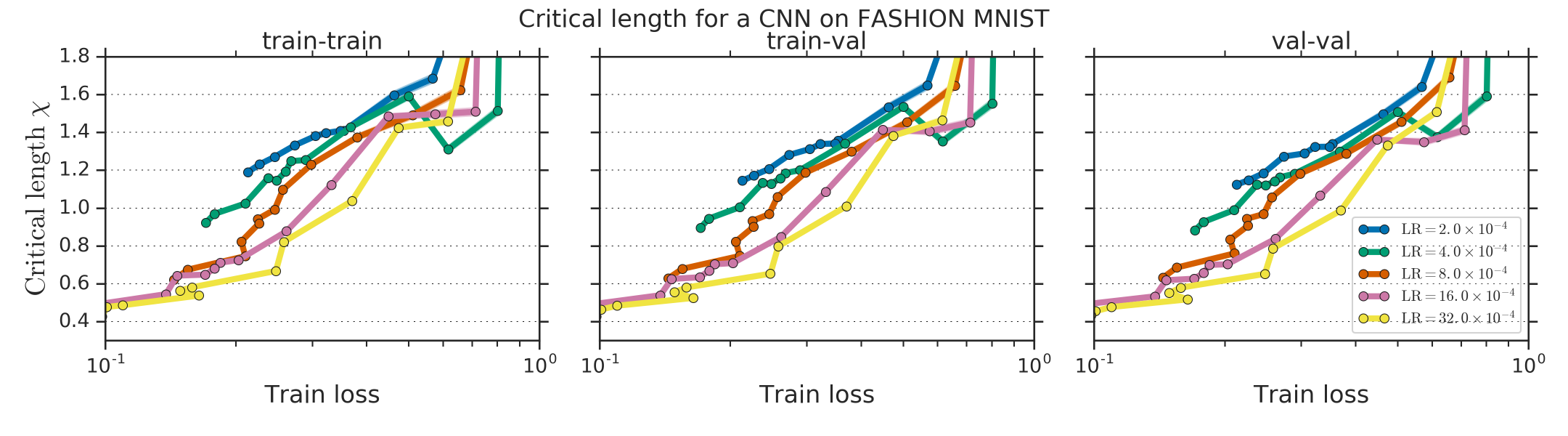} 
		\includegraphics[width=1.0\linewidth]{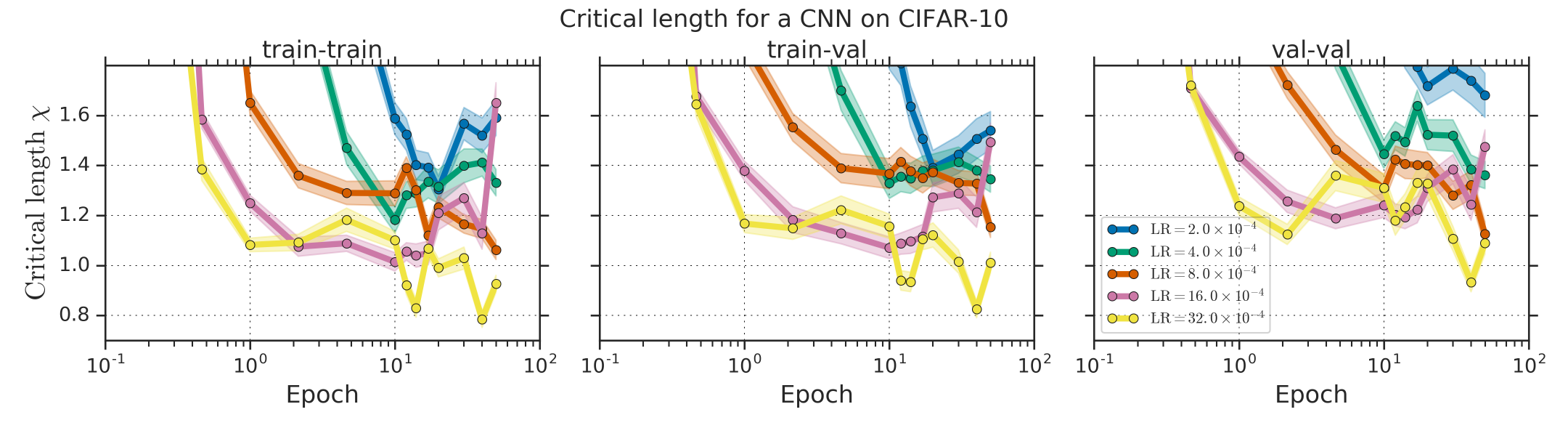} 
		\includegraphics[width=1.0\linewidth]{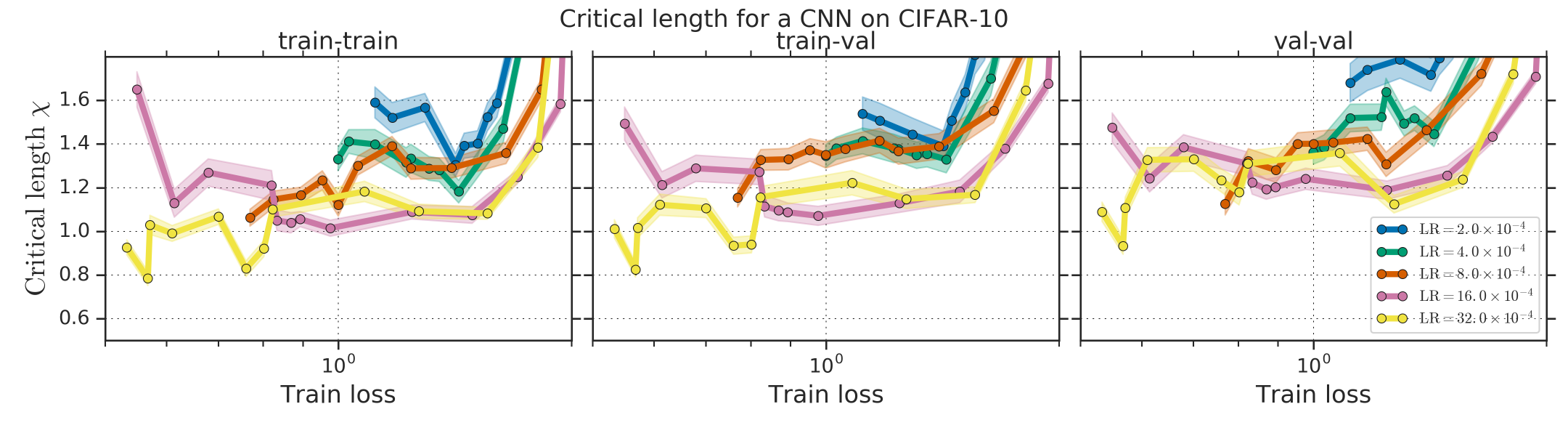}
		\caption{The effect of learning rate on the stiffness dynamical critical length $\xi$ -- the input space distance between images over which stiffness disappears (see Figure~\ref{fig:3D_diagram} for a cartoon illustration). The upper part shows $\xi$ for 5 different learning rates as a function of epoch, while the lower part shows it as a function of training loss in order to be able to compare different learning rates fairly. The larger the learning rate, the smaller the stiff domains, even given the same training loss. This means that the learning rate affects the character the function learned beyond the speed of training.}
		\label{fig:domain_size}
	\end{figure}
	At each epoch of training for each of our experiments, we analyzed the distribution of within-class stiffness between images based on their distance, and extracted the zero crossing which we call the critical dynamical length $\xi$. In Figures~\ref{fig:domain_size} and in the Supplementary material we summarize the dependence of $\xi$ on the epoch of training as well as the training loss for 5 different learning rates. We use the training loss to make sure we are comparing runs with different learning rates at the equivalent stages of training. We see that the bigger the learning rate, the smaller the domain size $\xi$. This is a surprising result -- higher learning rates doe not only lead to a faster training, they in fact influence properties of the learning function beyond that. We observe that higher learning rates lead to functions of lower $\xi$, i.e. functions that seem easier to \textit{bend} using gradient updates.  
	
	\section{Discussion}
	\label{sec:discussion}
	We explored the concept of neural network stiffness and used it to diagnose and characterize generalization. We studied stiffness for models trained on real datasets, and measured its variation with training iteration, class membership, distance between data points, and the choice of learning rate. We explored stiffness between pairs of data points coming both from the training set, one from the training and one from the validation set, and both from the validation set. Training-validation stiffness is directly related to the transfer of improvements on the training set to the validation set. We used two different stiffness metrics -- the sign stiffness and the cosine stiffness -- to highlight different phenomena.

	We explored models trained vision tasks (MNIST, FASHION MNIST, and CIFAR-10/100) and an natural language task (MNLI) through the lens of stiffness. In essence, stiffness measures the alignment of gradients taken at different input data points, which we show is equivalent to asking whether a weight update based on one input will benefit the loss on another. We demonstrate the connection between stiffness and generalization and show that with the onset of overfitting to the training data, stiffness decreases and eventually reaches 0, where even gradient updates taken with respect images of a particular class stop benefiting other members of the same class. This happens even within the training set itself, and therefore could potentially be used as a diagnostic for early stopping to prevent overfitting.
	
	Having established the usefulness of stiffness as a diagnostic tool for generalization, we explored its dependence on class membership. We find that in general gradient updates with respect to a member of a class help to improve loss on data points in the same class, i.e. that members of the same class have high stiffness with respect to each other. This holds at initialization as well as throughout most of the training. The pattern breaks when the model starts overfitting to the training set, after which within-class stiffness eventually reaches 0. We observe this behavior with fully-connected and convolutional neural networks on MNIST, FASHION MNIST, and CIFAR-10/100. 
	
	Stiffness between inputs from different classes relates to the generality of the features being learned and within-task transfer of improvement from class to class. With the onset of overfitting, the stiffness between different classes regresses to 0, as does within-class stiffness. We extended this analysis to groups of classes (=super-classes) on CIFAR-100, and went a step above that to super-super-classes (groups of super-classes). Using these semantically meaningful ways of grouping images together, we verified that the network is aware of more coarse-grained semantic groups of images, going beyond the 100 fine-grained classes it was trained on, and verified that stiffness as a useful tool to diagnose that.
	
	We also investigated the characteristic size of stiff regions in our trained networks at different stages of training (see Figure~\ref{fig:3D_diagram} for an illustration). By studying stiffness between two datapoints and measuring their distance in the input space, we observed that the farther the datapoints and the higher the epoch of training, the less stiffness there exists between them on average. This allowed us to define the \textit{dynamical critical length} $\xi$ -- an input space distance over which stiffness between input points decays to 0. $\xi$ corresponds to the size of stiff regions -- patches of the data space that can move together when a gradient update is applied, provided it were an infinitesimal gradient step.
	
	We investigated the effect of learning rate on stiffness by observing how $\xi$ changes as a function of epoch and the training loss for different learning rates. We show that the higher the learning rate, the smaller the $\xi$, i.e. for high learning rates the patches of input space that are improved together are smaller. This holds both as a function of epoch and training loss, which we used in order to compare runs with different learning rates fairly. This points towards the regularization role of learning rate on the kind of function we learn. We observe significant differences in the characteristic size of regions of input space that react jointly to gradient updates based on the learning rate used to train them.
	
\section{Conclusion}
\label{sec:conclusion}

\textbf{Why stiffness?} We find stiffness to be a concept worthwhile of study since it bridges several different, seemingly unrelated directions of research. 1) It directly relates to generalization, which is one of the key questions of the field, 2) it is (as we have verified) sensitive to semantic content of the input datapoints beyond the level of classes on which the neural net was trained explicitly, 3) it relates to the Hessian of the loss with respect to the weights, and therefore to the loss landscape and optimization literature. Understanding its properties therefore holds a potential for linking these areas together. 

\textbf{Empirical observations.} In this paper, we have observed the dependence of stiffness on a) class membership, b) distance between datapoints, c) learning rate, and d) stage of training. We did this on a range of vision tasks and a natural language task. We observed that stiffness is sensitive to the semantic content of images going beyond class, to super-class, and even super-super-class level on CIFAR-100. By defining the concept of the dynamical critical length $\xi$, we showed that higher learning rates lead to learned functions that are easier to \textit{bend} (i.e. they are less stiff) using gradient updates even though they perform equally well in terms of accuracy.  
	
\textbf{Future directions.} One immediate extension to the concept of stiffness would be to ascertain the role stiffness might play in architecture search. For instance, we expect locality (as in CNN) to reflect in higher stiffness properties. It is quite possible that stiffness could be a guiding parameter for meta-learning and explorations in the space of architectures, however, this is beyond the scope of this paper and a potential avenue for future work.

	\clearpage
	
	\bibliography{stiffness_bibliography}
	\bibliographystyle{unsrtnat}


	\clearpage
	
	\clearpage\newpage
	\appendix
	{\begin{center} {\Large{\textbf{Supplementary Material}}}
		\end{center}}
		\setcounter{figure}{0}
		\setcounter{table}{0}
		\makeatletter 
		\renewcommand{\thefigure}{S\@arabic\c@figure}
		\renewcommand{\thetable}{S\@arabic\c@table}
		\makeatother

		\begin{figure}[htbp]
			\centering
			\includegraphics[width=1.0\linewidth]{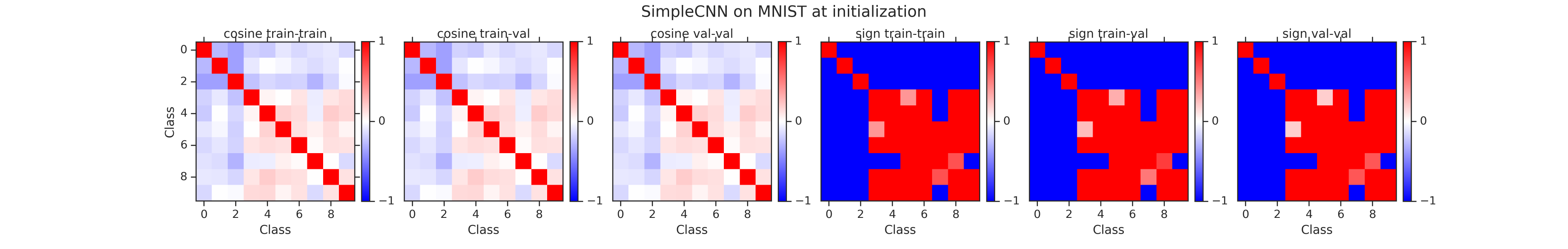}
			\includegraphics[width=1.0\linewidth]{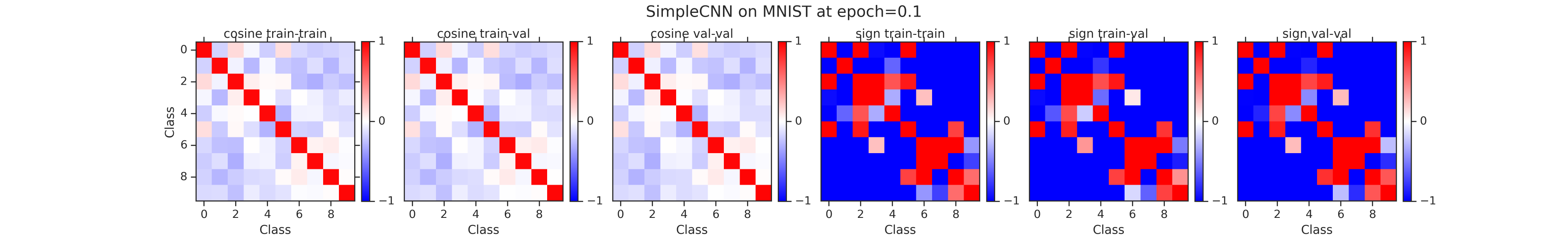}
			\includegraphics[width=1.0\linewidth]{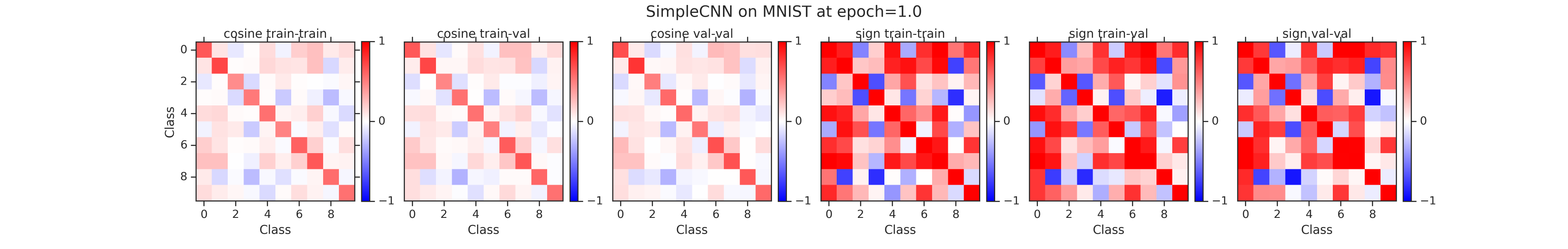}
			\includegraphics[width=1.0\linewidth]{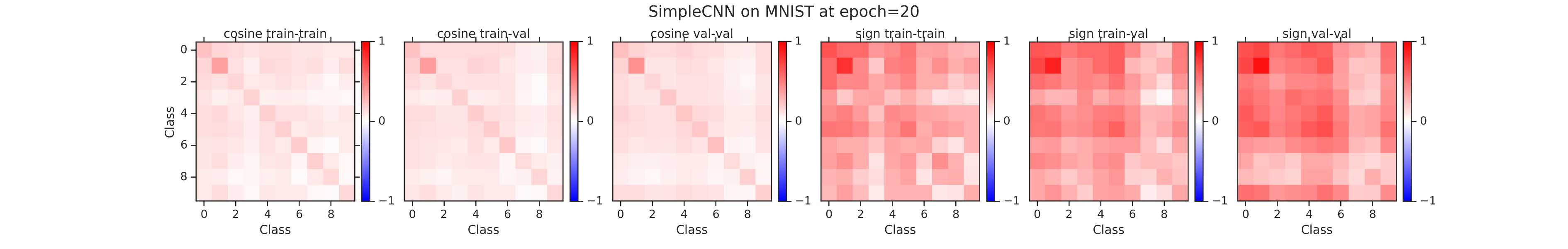}
			\caption{Class-membership dependence of stiffness for a CNN on MNIST at 4 different stages of training.}
			\label{fig:class_stiffness_CNN_MNIST}
		\end{figure}

		\begin{figure}[htbp]
			\centering
			\includegraphics[width=1.0\linewidth]{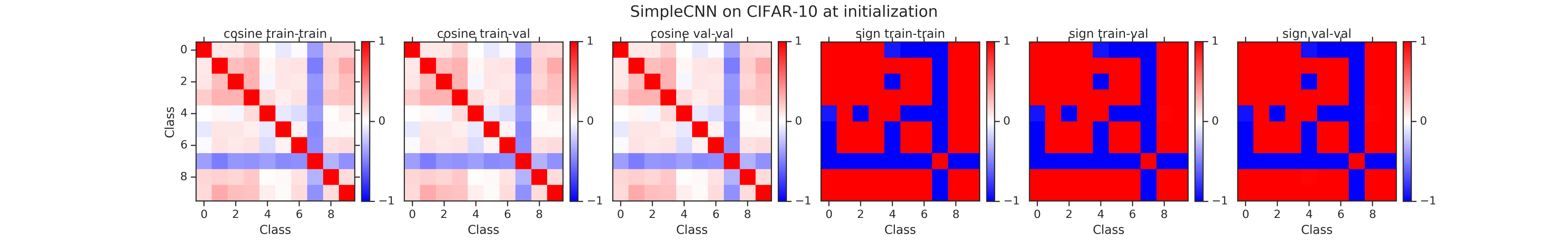}
			\includegraphics[width=1.0\linewidth]{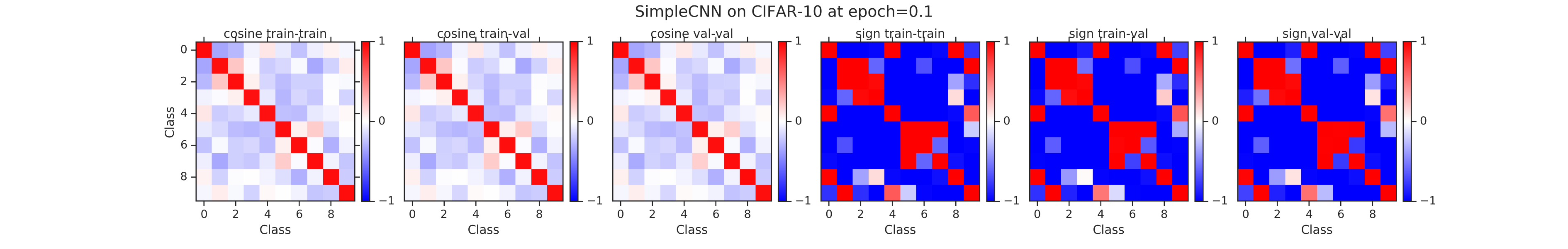}
			\includegraphics[width=1.0\linewidth]{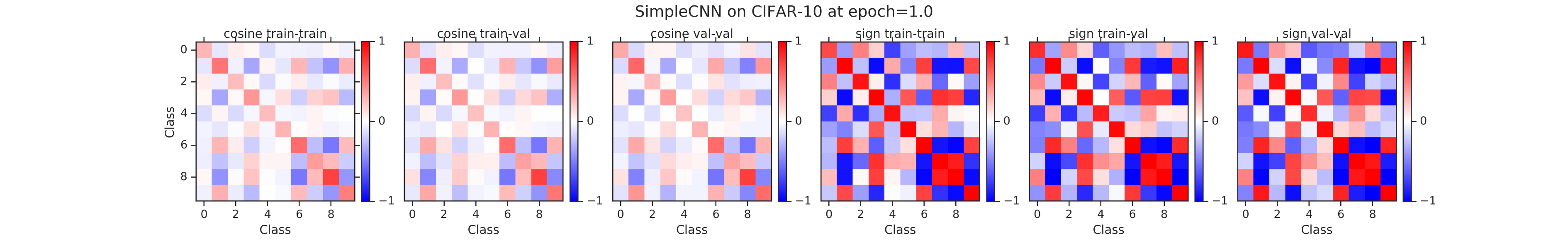}
			\includegraphics[width=1.0\linewidth]{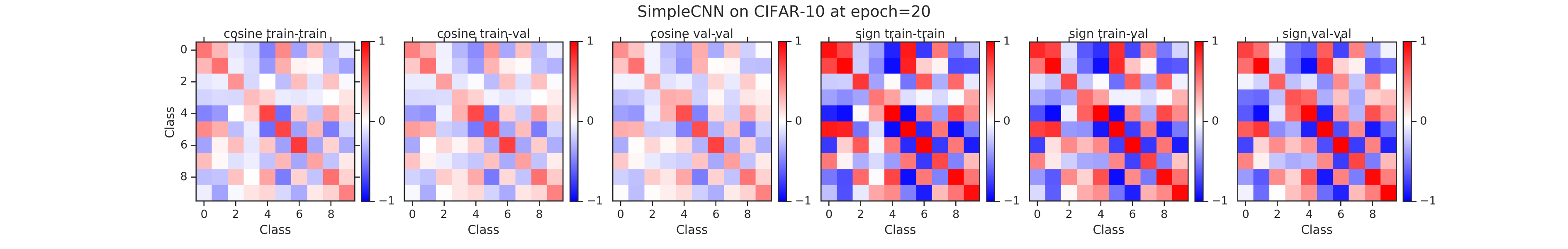}
			\caption{Class-membership dependence of stiffness for a CNN on CIFAR-10 at 4 different stages of training.}
			\label{fig:class_stiffness_CNN_CIFAR10}
		\end{figure}
		\begin{figure}[!h]
			\centering
			\includegraphics[width=1.0\linewidth]{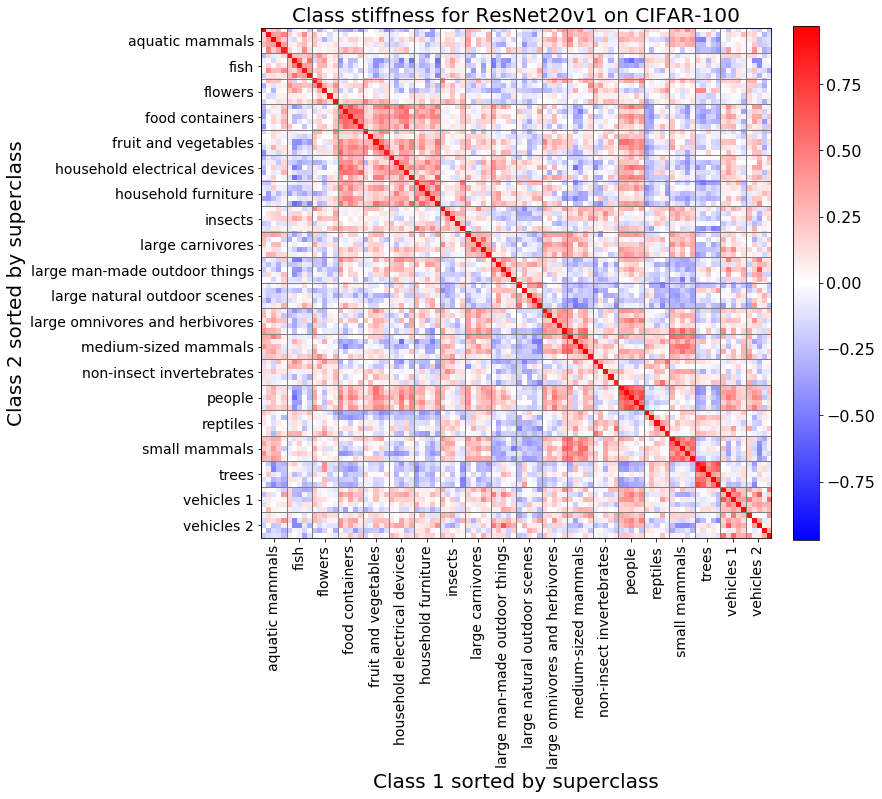}
			\caption{Stiffness for ResNet20v1 trained on CIFAR-100 at test accuracy 0.02.}
			\label{fig:CIFAR100_ResNet_acc0d02}
		\end{figure}
		\begin{figure}[!h]
			\centering
			\includegraphics[width=1.0\linewidth]{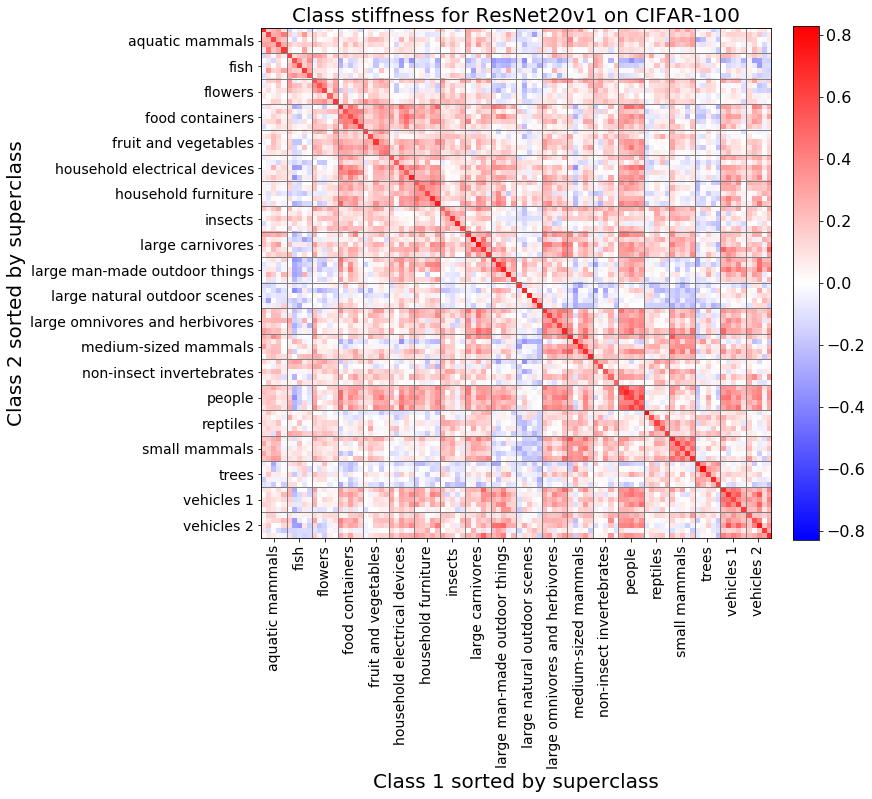}
			\caption{Stiffness for ResNet20v1 trained on CIFAR-100 at test accuracy 0.05.}
			\label{fig:CIFAR100_ResNet_acc0d02}
		\end{figure}
		\begin{figure}[!h]
			\centering
			\includegraphics[width=1.0\linewidth]{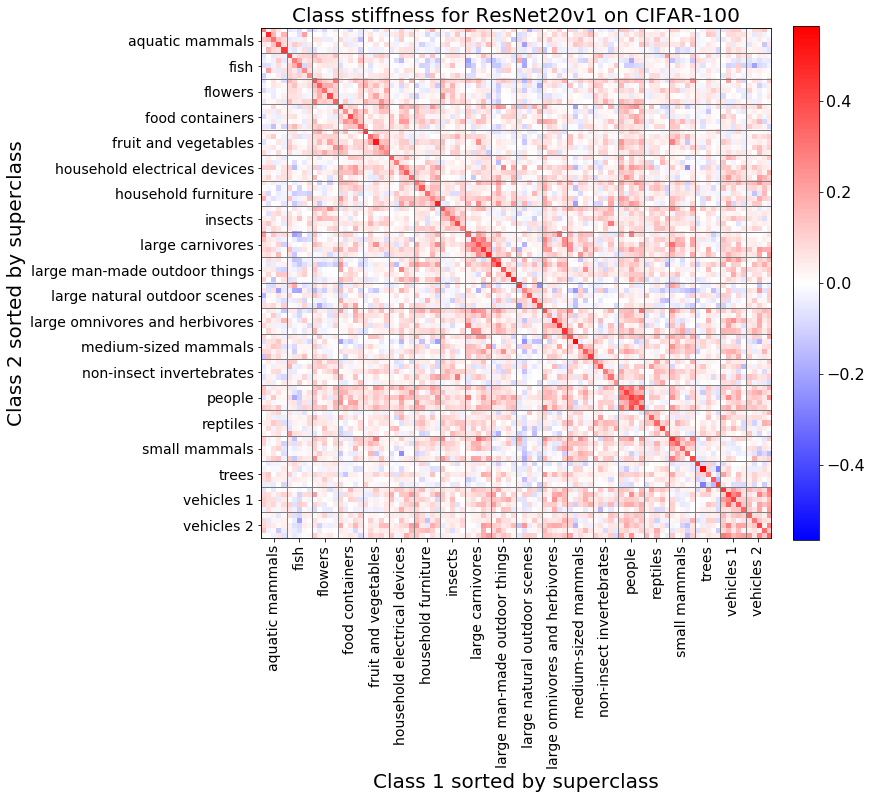}
			\caption{Stiffness for ResNet20v1 trained on CIFAR-100 at test accuracy 0.35.}
			\label{fig:CIFAR100_ResNet_acc0d02}
		\end{figure}
		\begin{figure}[!h]
			\centering
			\includegraphics[width=1.0\linewidth]{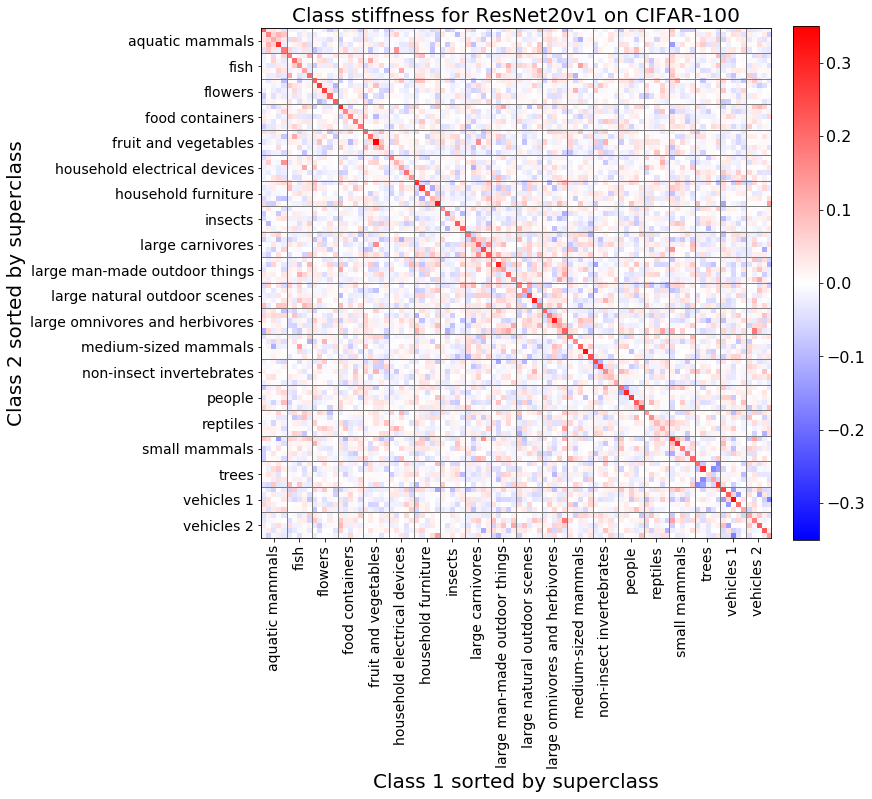}
			\caption{Stiffness for ResNet20v1 trained on CIFAR-100 at test accuracy 0.60.}
			\label{fig:CIFAR100_ResNet_acc0d02}
		\end{figure}
		
		\begin{figure}[!h]
			\centering
			\includegraphics[width=0.48\linewidth]{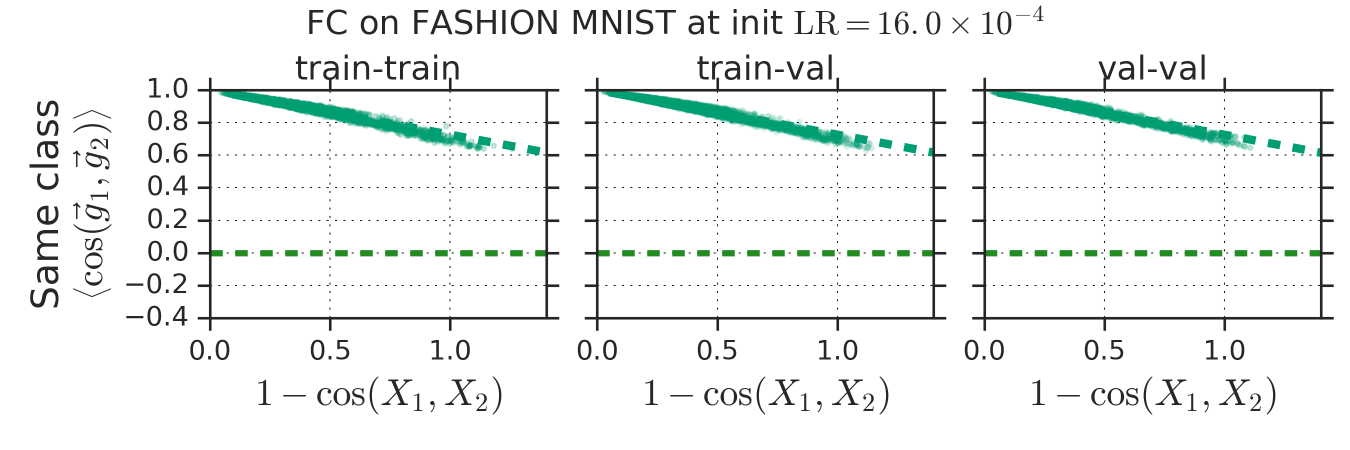}
			\includegraphics[width=0.48\linewidth]{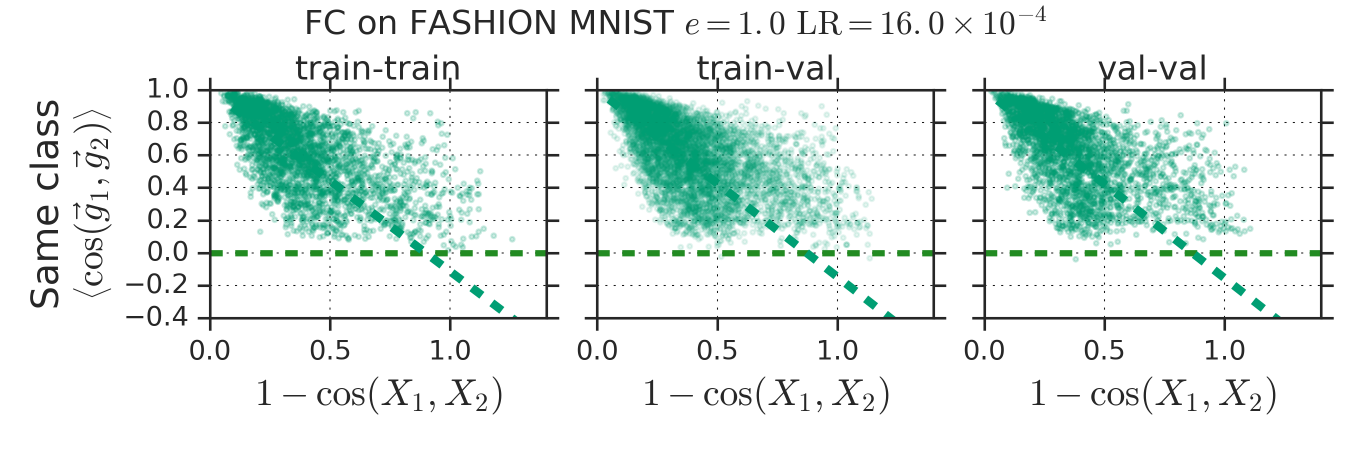}
			\\
			\includegraphics[width=0.48\linewidth]{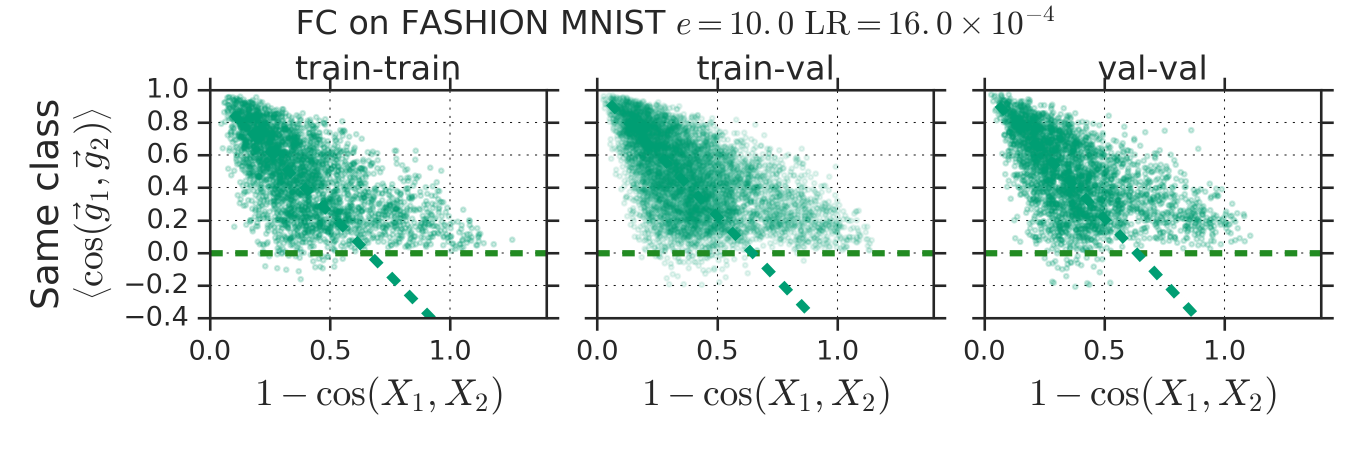}
			\includegraphics[width=0.48\linewidth]{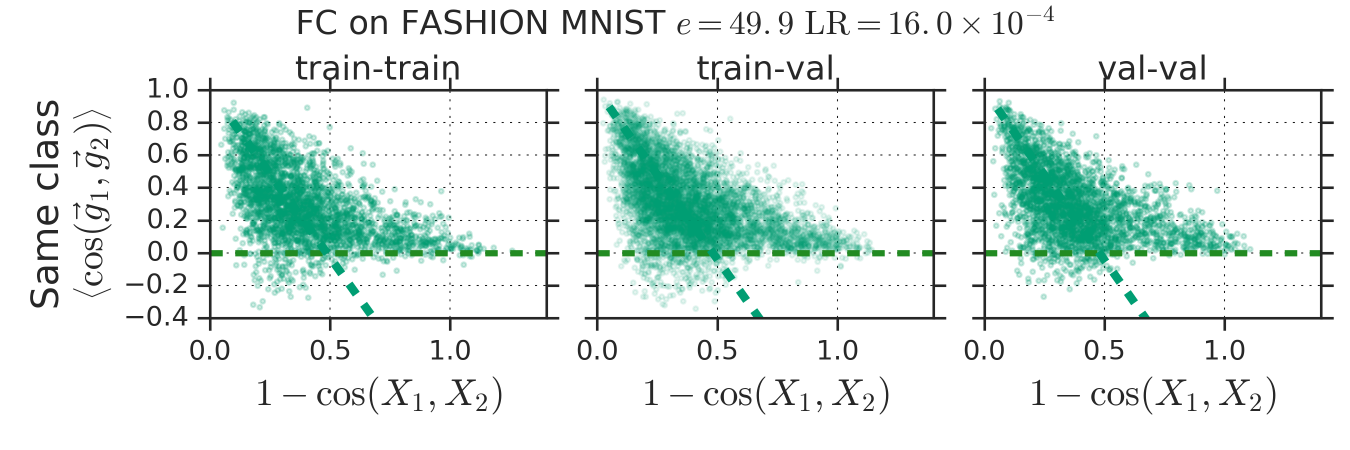}
			\caption{Stiffness between images of the same class as a function of their input space distance for 4 different stages of training of a fully-connected network (FC) on FASHION MNIST.}
			\label{fig:distance_dependence_FC_FASHIONMNIST}
		\end{figure}
		\begin{figure}[!h]
			\centering
			\includegraphics[width=0.48\linewidth]{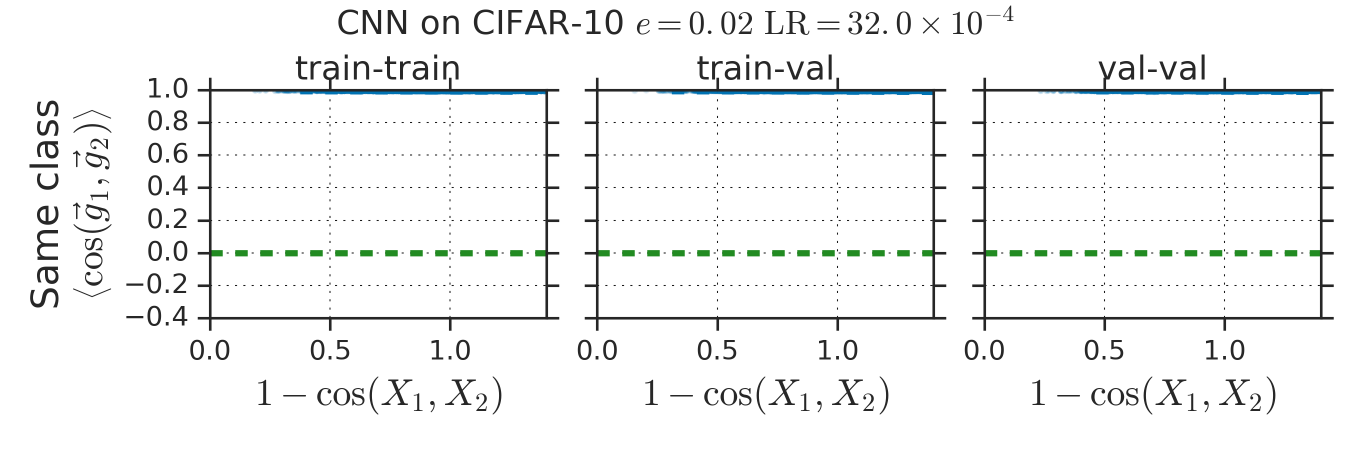}
			\includegraphics[width=0.48\linewidth]{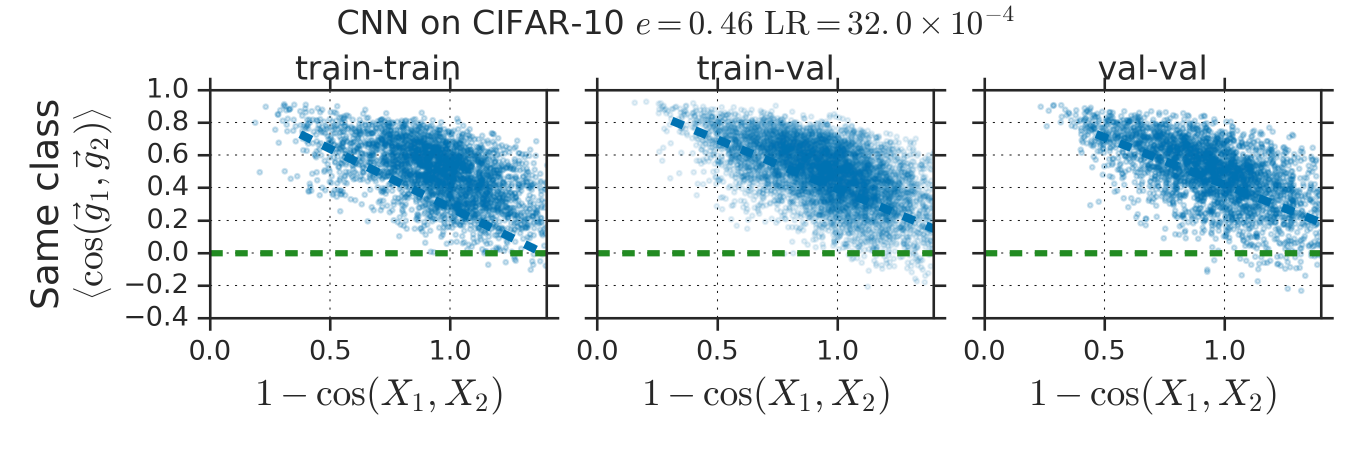}
			\\
			\includegraphics[width=0.48\linewidth]{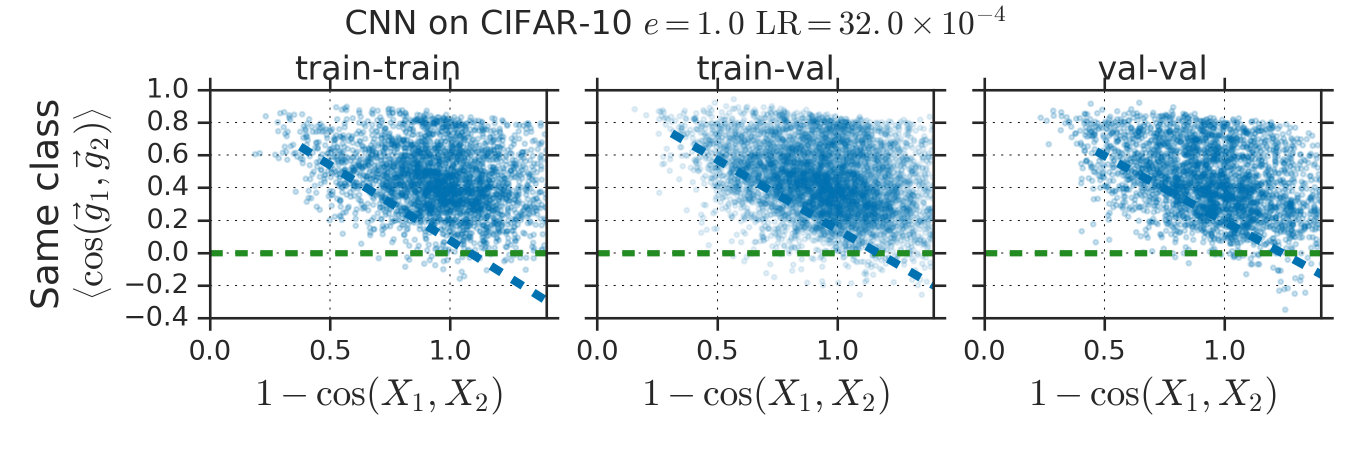}
			\includegraphics[width=0.48\linewidth]{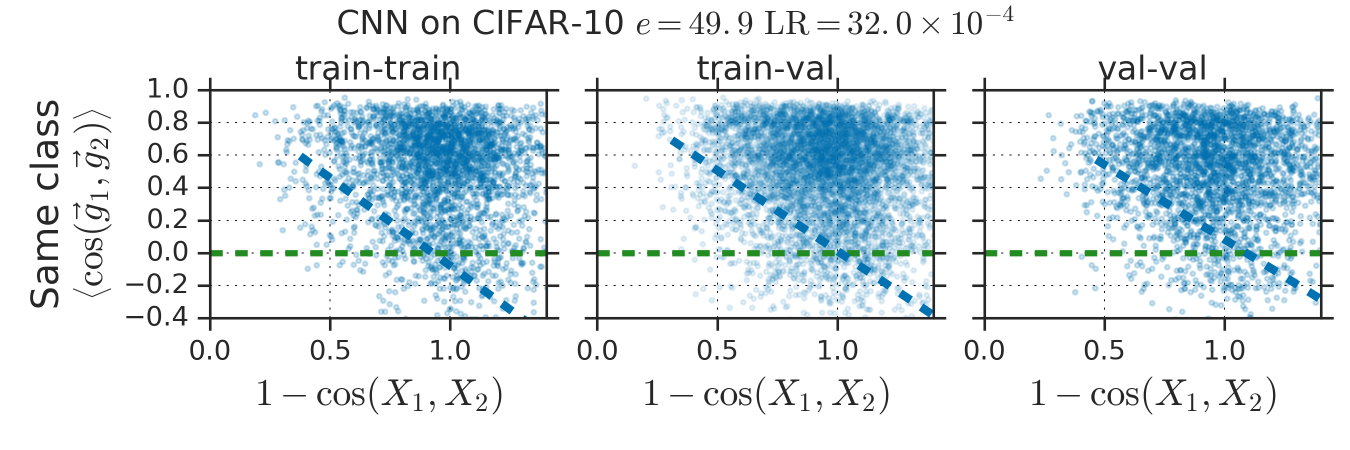}
			\caption{Stiffness between images of the same class as a function of their input space distance for 4 different stages of training of a CNN on CIFAR-10.}
			\label{fig:distance_dependence_CNN_CIFAR10}
		\end{figure}
		\begin{figure}[!h]
			\centering
			\includegraphics[width=1.0\linewidth]{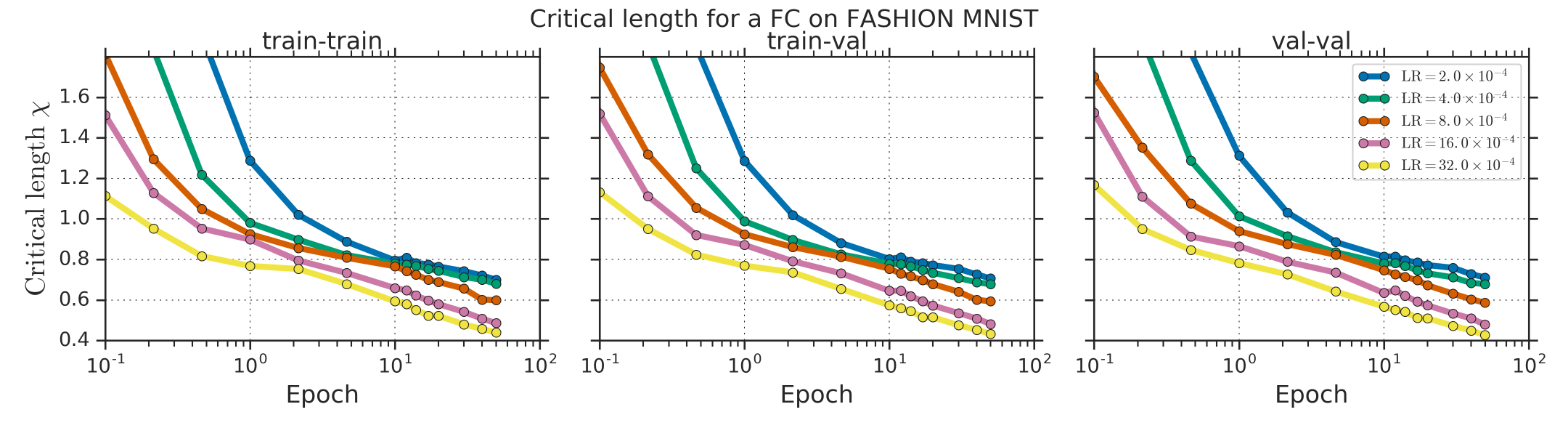} 
			\includegraphics[width=1.0\linewidth]{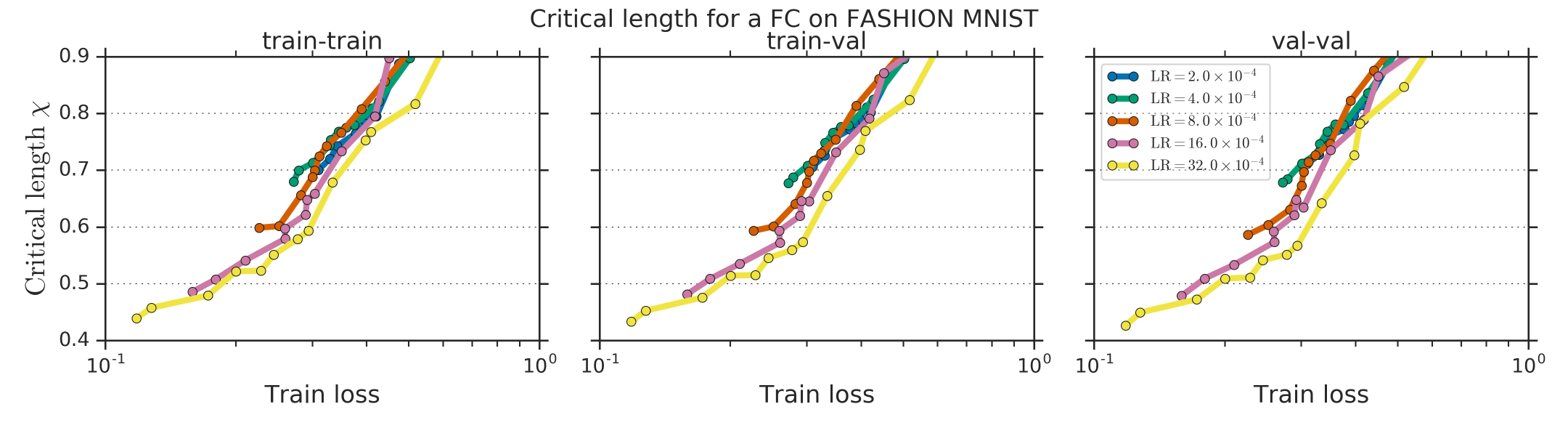}
			\caption{The effect of learning rate on the stiffness dynamical critical length $\xi$ -- the input space distance between images over which stiffness disappears. The upper part shows $\xi$ for 5 different learning rates as a function of epoch, while the lower part shows it as a function of training loss in order to be able to compare different learning rates fairly. The larger the learning rate, the smaller the stiff domains.}
			\label{fig:domain_size_additional}
		\end{figure}

\end{document}